\documentclass[sigconf]{acmart}
\usepackage{multirow}
\usepackage[linesnumbered,ruled,boxed]{algorithm2e} 

\usepackage{amssymb}
\usepackage{amsmath}
\usepackage{booktabs}
\usepackage{colortbl}
\newcommand{\RNum}[1]{\uppercase\expandafter{\romannumeral #1\relax}}
\AtBeginDocument{%
	\providecommand\BibTeX{{%
			\normalfont B\kern-0.5em{\scshape i\kern-0.25em b}\kern-0.8em\TeX}}}

\setcopyright{acmcopyright}
\copyrightyear{2022}
\acmYear{2022}
\setcopyright{acmcopyright}
\acmConference[SIGIR '22]{Proceedings of the 45th International ACM SIGIR 
Conference on Research and Development in Information Retrieval}{July 11--15, 
2022}{Madrid, Spain.}
\acmBooktitle{Proceedings of the 45th International ACM SIGIR Conference on 
Research and Development in Information Retrieval (SIGIR '22), July 11--15, 
2022, Madrid, Spain}
\acmPrice{15.00}
\acmISBN{978-1-4503-8732-3/22/07}
\acmDOI{10.1145/3477495.3532016}

\begin{document}
\fancyhead{}

\title{Logiformer: A Two-Branch Graph Transformer Network \\ for Interpretable Logical Reasoning}

\author{Fangzhi Xu}
\affiliation{%
	\institution{School of Computer Science and Technology, Xi'an Jiaotong 
	University}
	\city{Xi'an}
	\country{China}}
\email{Leo981106@stu.xjtu.edu.cn}

\author{Jun Liu}
\authornote{The corresponding author.}
\affiliation{%
	\institution{Shaanxi Province Key Laboratory of Satellite and Terrestrial 
	Network Tech. R\&D, National Engineering lab for Big Data Analytics}
	\city{Xi'an}
	\country{China}}
\email{liukeen@xjtu.edu.cn}

\author{Qika Lin}
\affiliation{%
	\institution{School of Computer Science and Technology, Xi'an Jiaotong 
	University}
	\city{Xi'an}
	\country{China}}
\email{qikalin@foxmail.com}

\author{Yudai Pan}
\affiliation{%
	\institution{School of Computer Science and Technology, Xi'an Jiaotong 
		University}
	\city{Xi'an}
	\country{China}}
\email{pyd418@foxmail.com}

\author{Lingling Zhang}
\affiliation{%
	\institution{School of Computer Science and Technology, Xi'an Jiaotong 
		University}
	\city{Xi'an}
	\country{China}}
\email{zhanglling@xjtu.edu.cn}

\begin{abstract}
Machine reading comprehension has aroused wide concerns, since it explores the 
potential of model for text understanding. To further equip the machine with 
the reasoning capability, the challenging task of logical reasoning is 
proposed. Previous works on logical reasoning have proposed some strategies to 
extract the logical units from different aspects. However, there still remains 
a challenge to model the long distance dependency among the logical units. 
Also, it is demanding to uncover the logical structures of the text and further 
fuse the discrete logic to the continuous text embedding. To tackle the above 
issues, we propose an end-to-end model Logiformer which utilizes a two-branch 
graph transformer network for logical reasoning of text. Firstly, we introduce 
different extraction strategies to split the text into two sets of logical 
units, and construct the logical graph and the syntax graph respectively. The 
logical graph models the causal relations for the logical branch while the 
syntax graph captures the co-occurrence relations for the syntax branch. 
Secondly, to model the long distance dependency, the node sequence from each 
graph is fed into the fully connected graph transformer structures. The two 
adjacent matrices are viewed as the attention biases for the graph transformer 
layers, which map the discrete logical structures to the continuous text 
embedding space. Thirdly, a dynamic gate mechanism and a question-aware 
self-attention module are introduced before the answer prediction to update the 
features. The reasoning process provides the interpretability by employing the 
logical units, which are consistent with human cognition. The experimental 
results show the superiority of our model, which outperforms the 
state-of-the-art single model on two logical reasoning benchmarks. 
\footnote{The code is public in \url{https://github.com/xufangzhi/Logiformer}.}
\end{abstract}



\maketitle

\section{Introduction}
\begin{figure}[t]
	\large
	\centering
	\includegraphics[scale=0.50]{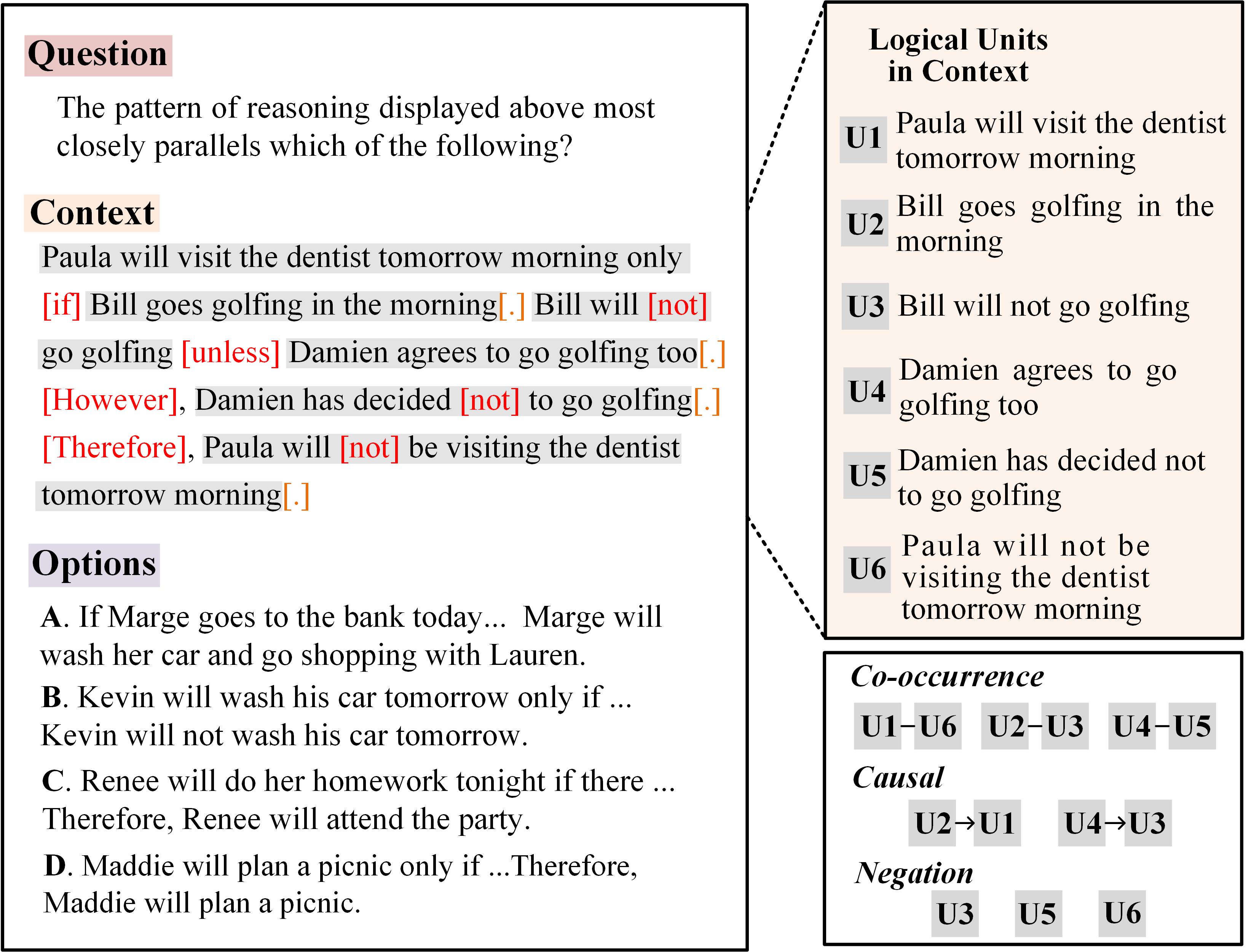}
	\caption{An example of the logical reasoning task and some detailed illustrations.}
	\label{fig_dataset}
	\vspace{-0.4cm}
\end{figure}

\noindent Machine reading comprehension \cite{liu2019neural, zhang2019machine, 
hirschman2001natural} has been one of the major focuses in the field of Natural 
Language Processing (NLP) \cite{chowdhury2003natural, hirschberg2015advances} 
in recent years. A large number of models have achieved competitive 
performances in some famous datasets, such as SQuAD\cite{rajpurkar2016squad, 
rajpurkar2018know}, RACE\cite{lai2017race}. However, these models 
\cite{seo2016bidirectional, dhingra2016gated, yu2018qanet} lack the capability 
of logical reasoning. To facilitate the machine for human intelligence, the 
task of logical reasoning MRC \cite{yu2019reclor,liu2020logiqa} was proposed 
previously. Similar to the traditional MRC, the task of logical reasoning also 
requires the models to predict the answers depending on the given text inputs. 
Figure 1 illustrates an logical reasoning example from ReClor dataset 
\cite{yu2019reclor}. The inputs include the context, question and a set of 
options. One of the unique characteristics of the text is the rich logical 
structures. As illustrated in Figure \ref{fig_dataset}, the logical structure 
of the context can be uncovered in a certain way. We define the split short 
sentences as logical units (e.g., U1-U6). The logical units contain the 
independent and complete semantics, which are not kept in the token-level text 
features. The understanding of the text requires the global semantics of each 
logical unit, as well as the interactions among them based on some logical 
relations (e.g., \textit{causal} and \textit{co-occurrence}). Therefore, the 
main challenges for the task of logical reasoning can be summarized as the 
following two aspects.

Firstly, it remains a challenge to model the long distance dependency 
\cite{hockett1947problems} of the extracted logical units. Some previous 
methods, such as DAGN \cite{huang2021dagn}, have proposed to split the text 
into discourse nodes \cite{xu2019discourse} and constructed a sequential chain 
graph for reasoning. However, it neglects the natural long-distance dependency 
among logical units. For example in Figure 1, the first and the last sentences 
share the same subject (\textit{Paula}) and predicate (\textit{visit the 
dentist}), though they are distant in the graph space. The chain structure 
limits the information update. In a word, the simple graph structure built for 
the logical text would fail to provide the efficient one-hop interaction 
\cite{salha2020simple}. Pretrain-based transformer structures 
\cite{vaswani2017attention} have the natural advantage of modeling the long 
text and show excellent performance on the popular tasks. To enhance the 
logical perception ability of the language models, previous works have 
attempted to employ additional segment embedding at the beginning. However, it 
is still limited to the token-level interactions, which sacrifices the global 
semantics of logical units. Take the first sentence of the context in Figure 1 
as an instance, two extracted units of \textit{Paula will visit the dentist 
tomorrow morning}(U1) and \textit{Bill goes golfing in the morning}(U2) express 
the causal relations within this sentence. The token-aware models would stress 
more on the text semantics and fail to capture such logical information.

Secondly, it is intractable to bridge the gap between the discrete logical 
structures and the continuous text embedding space. Take a closer observation 
of the logical units in Figure 1, the units are not completely separated. In 
this work, we pay much attention to the two explicit relationships 
(\textit{causal} and \textit{co-occurrence}). We summarize them into the 
logical branch and syntax branch respectively. From the logical branch, 
connectives (e.g., \textit{if}, \textit{unless}, \textit{Therefore}) play a 
significant role in covering the logical relations. For example in Figure 
\ref{fig_dataset}, the logical units U1 and U2 are connected with the 
connective \textit{if} while U3 and U4 share the connective \textit{unless}. 
From the syntax branch, the logical units are not independent but have some 
repeating occurrences. For example, the units of \textit{Bill goes golfing in 
the morning} and \textit{Bill will not go golfing} belong to the co-occurrence 
information, which have strong correlations. Thus, these syntactic 
relationships help make the corresponding logical units closer in structure. 
The above connection structures from two branches are discrete, which is 
incompatible to the continuous text representation. Some early works simply 
feed the text to the Pretrained Language Models (PLMs) \cite{yang2019xlnet, 
clark2020electra} and rely on the context to learn the logical semantics. But 
it includes much noise in the text embedding and lacks the potential of 
interpretability. Previously, LReasoner \cite{wang2021logic} proposes a method 
to transform the logical expressions to text based on the templates and feeds 
the extended text into PLM. However, it still embeds the logic in an implicit 
form and fails to make up for the weakness of PLMs in logical reasoning. Some 
works like FocalReasoner \cite{ouyang2021fact} uncovers the logical structure 
in only one aspect (e.g., capture the co-occurrence between units). It leads to 
the weak capability of the model to capture logical relationships.

In light of the above challenges, we propose a novel model named 
\textbf{Logiformer} which is an end-to-end architecture with graph transformer 
for interpretable logical reasoning of text. By employing the fully connected 
graph transformer structure to enhance the direct interactions, we tackle the 
issue of long distance dependency among the logical units. To encode the 
discrete logical structures to the continuous text embedding space, we apply 
the attention biases from both the logical and syntax branches. The whole 
reasoning is on the basis of logical units and the built graphs, which are 
consistent with the human cognition. The explicit relations among units and the 
weighted attention maps provide the interpretability for the logical reasoning. 
In details, firstly, Logiformer split the text into logical units and construct 
the logical graph based on the causal relations for the logical branch. For the 
syntactic branch, the split nodes and a syntax graph are also obtained. 
Secondly, we feed the node sequences and two graph topology to the fully 
connected graph transformers respectively. The respective adjacent matrices are 
viewed as the attention biases to encode the logical structures to each graph 
transformer. Thirdly, we combine the updated features from two branches with a 
dynamic gate mechanism. With the additional token-level embedding, we can map 
the features to the same space. By means of the question-aware self-attention, 
the final feature can be utilized to predict the answers.

The main contributions are listed as follows:

\begin{itemize}
	\item A two-branch graph transformer network named Logiformer is proposed to model the long distance dependency of the logical units and encode the discrete logical structure to the continuous text embedding. As far as we know, we are the first to tackle both issues in the logical reasoning task.
	\item In light of drawbacks of chain-type text graphs, we take the fully connected structures into consideration, containing the awareness of both logic and syntax simultaneously. Two graphs are constructed based on the extracted logical units and their topology is utilized as the attention biases.
	\item The extraction of the logical units and the explicit relations are consistent with the human cognition. The uncovered logical structures and the weighted attention maps of the logical units provide the excellent interpretability for the logical reasoning process.
	\item Extensive experiments show that Logiformer outperforms the state-of-the-art (SOTA) results with single model on two logical reasoning datasets. Furthermore, ablation studies prove the effectiveness of each module in our model.
\end{itemize}

\section{Related Work}
In this section, we will introduce the current researches on MRC and logical reasoning.
\vspace{-0.2cm}
\subsection{Machine Reading Comprehension}
Recent years have witnessed the rapid growth of MRC\cite{liu2019neural}, where the model is required to infer the answers based on the given context and a question. A variety of datasets have been proposed to check the performances of MRC models. Among them, SQuAD\cite{rajpurkar2016squad, rajpurkar2018know} focuses on the span extractions on the factual questions. HotpotQA\cite{yang2018hotpotqa} and OpenBookQA\cite{mihaylov2018can} require the multi-hop reasoning capability of the models. A couple of multiple choice datasets like RACE\cite{lai2017race} cover the examinations for middle or high school students. Some representative models achieve great success on these datasets. Retro-Reader \cite{zhang2021retrospective} applies a two-state strategy to solve the questions. But it mainly investigates the overall interactions of the context and question, which fails to deal with the complex logic within the text. SG-Net \cite{zhang2020sg} integrates the syntax information into the self-attention module to improve the performance, but it does not show the potential on tackling the logical information. Generally speaking, the datasets mentioned above rely much on the token-level matching, which can be well tackled with large-scale pretraining models like BERT\cite{devlin2018bert} and GPT-3\cite{brown2020language}. To make the models closer to the human intelligence, it is necessary to introduce more challenging tasks requiring logical reasoning. Previously, the task of Natural Language Inference(NLI)\cite{bowman2015large, storks2019recent} is proposed to motivate the models to infer the relations(i.e., Contradiction, Entailment and Neutral) between two sentences. Nevertheless, it is limited by the fixed inputs and outputs and fails to extend the task to more complex settings.

\subsection{Logical Reasoning}
To improve the reasoning ability of the models, several datasets on multiple choice have been proposed previously. ReClor\cite{yu2019reclor}, which is extracted from standardized graduate admission examinations and law school admission test, has aroused wide concerns. For better evaluation, it separates the biased examples into EASY set and the challenging ones into HARD set. LogiQA \cite{liu2020logiqa} is also one of the representatives, which also aims to improve the logical reasoning capability. It is sourced from expert-written questions and covers multiple types of deductive reasoning. Experiments show that previous SOTA models on traditional MRC perform bad on the two datasets. Under such circumstances, some of the recent works attempt to enhance logical reasoning from different perspectives. DAGN\cite{huang2021dagn} proposes a reasoning network based on the discourse units extracted from the text. But it simply forms a chain-type discourse network and weakens the relations between two distant units. FocalReasoner \cite{ouyang2021fact} stresses that fact units in the form of subject-verb-object are significant for logical reasoning. It constructs a supergraph on top of the fact units and updates the node features relying on Graph Neural Network. However, it ignores the relation connectives from the text and lacked the logical modeling. LReasoner \cite{wang2021logic} focuses on capturing symbolic logic from the text and puts forward a context extension framework based on logical equivalence laws. However, it relies heavily on the language models for token-level embedding and neglects the sentence-level interactions.

\section{Methodology}
\begin{figure*}[t]
	\large
	\centering
	\includegraphics[scale=0.52]{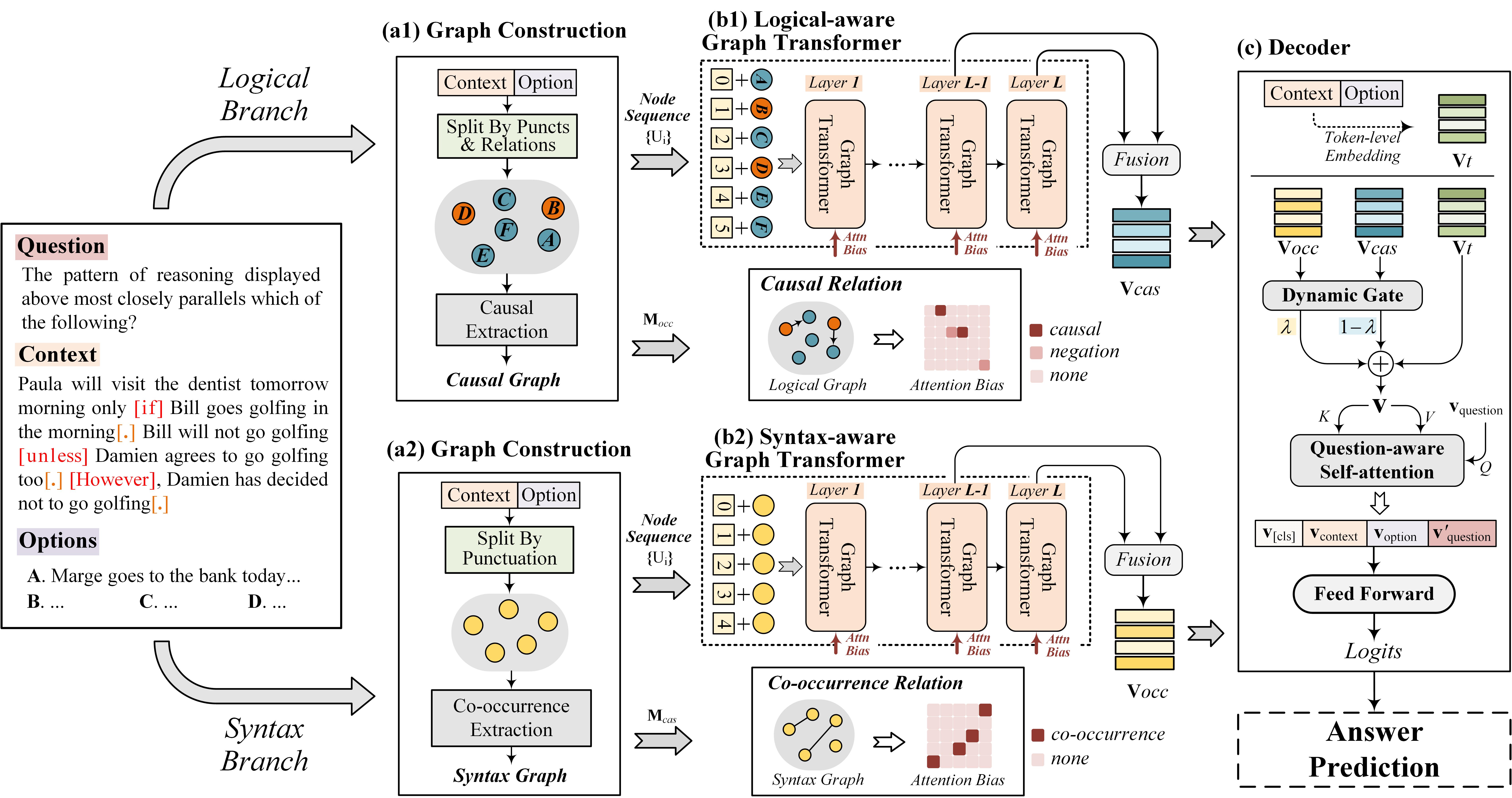}
	\caption{The architecture of Logiformer. The left part is an input example of the dataset. The graph construction modules (\textit{a1,a2}) split the text into logical units and build two graphs from two branches respectively. The graph transformer structures (\textit{b1,b2}) update the text features combined with the logical and syntactic relations. Finally, the decoder module (\textit{c}) is utilized to conduct the feature fusion and predict the answers.}
	\label{fig_model}
	\vspace{-0.1cm}
\end{figure*}

This section will introduce the proposed end-to-end model Logiformer. The architecture of Logiformer is shown in Figure \ref{fig_model}. The left part of the model is an example of the logical reasoning task. The understanding of text will be divided into two branches: logical branch (upper) and syntactic branch (lower). This architecture mainly includes the following three parts: a) graph construction from the text; b) logical-aware and syntax-aware graph transformers for feature updates; c) the decoder including a dynamic gate mechanism and a question-aware self-attention module.

\subsection{Task Formulation}
Given a dataset $\mathcal{D}$ for logical reasoning, which consists of $N$ examples totally. The inference of the $i^{th}$ question can be formulated as follows:

\vspace{-0.5cm}
\begin{equation}
	\hat{a} =\underset{a_{i,j} \in A_{i}}{\arg \max }  p\left(a_{i, j} \mid c_{i}, q_{i}, A_{i}; \theta\right),
\end{equation}
where $c_{i}$, $q_{i}$, $A_{i}$ represent the context, question sentence and candidate set respectively. The number of options in $A_{i}$ is $n$, $j \in [0,n-1]$ and $a_{i,j}\in A_{i}$ represents the $j^{th}$ option. $\hat{a}$ denotes the predicted option. $\theta$ denotes the trainable parameters.

Since the current methods mainly focus on the token-level representation of the text with the help of PLMs, they will naturally ignore some global semantics for each sentence or phrase. To capture the global feature within each sentence, we first obtain the text fragments, which are split by connectives or punctuations. We define the text fragment that reflect the complete semantic of an event or argument as the logical unit with the symbol $U$. Take the context in Figure \ref{fig_model} as an instance, we split the text by both connectives and punctuations and obtain a set of logical units shown in Table \ref{logical_units}.

\begin{table}[t]
	\centering
	\caption{The set of logical units from the example context (split by connectives and punctuations).}
	\vspace{-0.25cm}
	\begin{tabular}{cc}
		\toprule
		\textbf{Symbol} & \textbf{Logical Units} \\
		\hline
		$U_{1}$ & Paula will visit the dentist tomorrow morning\\
		$U_{2}$ & Bill goes golfing in the morning\\
		$U_{3}$ & Bill will not go golfing\\
		$U_{4}$ & Damien agrees to go golfing too\\
		$U_{5}$ & Damien has decided not to go golfing\\
		\bottomrule
	\end{tabular}
	\label{logical_units}
	\vspace{-0.4cm}
\end{table}

Considering that there exist explicit causal relations between units, we 
further introduce the conditional connective `$\to$'. And in some cases, it is 
required to reverse logical units for the negation expression, we also employ 
the operation `$\neg$'. Combining the logical units and causal connections in 
the form of conjunction, we can derive the logical expression of the text:

\vspace{-0.1cm}
\begin{equation}
	\left( {{U_2} \to {U_1}} \right) \wedge \left( {{U_4} \to {}^\neg {U_3}} \right) \wedge {}^\neg {U_5}.
\end{equation}

Obviously, there exist two key components in the logical expression: i) logical units $U_{k}$; ii) logical connectives, i.e., $\to$ and $\neg$. The former one focuses on the syntactic information, while the latter one is more related to logical structure of the context.

\subsection{Graph Construction}
Given the $i^{th}$ inputs, Logiformer first concatenates the context $c_{i}$ with each option $a_{i,j}$ respectively to form the input sequences. According to the previous analysis, Logiformer will tackle the inputs from two branches (i.e., logical branch and syntax branch) and build two graphs (i.e., logic graph and syntax graph) respectively.

\subsubsection{\textbf{Logical Graph}}
For the logical branch, Logiformer mainly concentrates on the causal relations. 
Considering that the causal relation often appears with explicit logical words 
such as \textit{if}, \textit{unless}, \textit{because}, we can leverage the 
explicit logical words as the basis of split. Therefore, we include 100 
commonly used logical words according to PDTB 2.0 \cite{prasad2008penn}.

Combining the explicit logical words and punctuations, we can separate the text 
sequence into logical units. Each unit serves as a node for future updates. 
Especially, we pick out the nodes pairs connected by the explicit causal 
relation words and name them as $condition$ $node$ (orange nodes) and $result$ 
$node$ (blue nodes). Meanwhile, we classify the common nodes which do not 
contain causal relations into $result$ $node$ (blue nodes). Thus, we obtain the 
node set from the perspective of logic.

According to the extracted causal node pairs, we can create directed connection 
from each condition node $p$ to result node $q$. This kind of connection is 
reflected in the adjacent matrix $\mathbf{M}_{cas} \in 
\mathbb{R}^{K_{cas}\times K_{cas}}$ of the logical graph as 
$\mathbf{M}_{cas}[p-1, q-1] = 1$.

Also, to avoid the semantic reverse brought by the negation, we mark the nodes with the explicit negation words (e.g., \textit{not}, \textit{no}). The node $k$ with negation semantics are expressed in the adjacent matrix as $\mathbf{M}_{cas}[k-1, k-1] = -1$.

Therefore, the logical graph has the perception of the causal relations and 
negations. And the obtained adjacent matrix $\mathbf{M}_{cas} \in 
\mathbb{R}^{K_{cas}\times K_{cas}}$ of the logical graph is asymmetric. 

\vspace{-0.25cm}
\subsubsection{\textbf{Syntax Graph}}
The main purpose of the syntactic understanding is to capture the inner relations between the logical units $U_{k}$. Noticing that some logical units share the common words or phrases in Figure \ref{fig_model}, e.g., \textit{Bill}, \textit{Damien} and \textit{go golfing}. It illustrates that the text has a strong characteristics of co-occurrence. Also, co-occurrence usually exists between two complete sentences. Therefore, we consider to split the text sequence only by punctuations and obtain a set of sentence nodes with no original connection. It is required to extract the co-occurrence between the sentence nodes. As each node consists of its related tokens, we propose a simple strategy to capture the co-occurrence, shown in Algorithm \ref{algorithm}.

\begin{algorithm}[t]
	\KwIn{Sentence Nodes $U_{k}$ ($k\in \{1,2,...,K_{occ}\}$), hyper-parameter $\delta$, stop words corpus $C_{s}$}
	\KwOut{Co-occurrence Matrix $\mathbf{M}_{occ}$}
	Initialize co-occurrence matrix $\mathbf{M}_{occ} \in \mathbb{R}^{K_{occ} \times K_{occ}} $ to zero.  \\
	\For{$k = 1,2,...,K_{occ}$}
	{
		Include all the tokens of $U_{k}$ into $Set_{k}$. Meanwhile, exclude stop words from $Set_{k}$ based on $C_{s}$ \\
		\For{$j = k+1,k+2,...,K_{occ}$}
		{
			Include all the tokens of $U_{j}$ into $Set_{j}$ and exclude stop words from $Set_{j}$ based on $C_{s}$ \\
			\tcc{Ensure the two sets not empty}
			\If{$len(Set_{k}) > 0$ and $len(Set_{j}) > 0$}
			{
				$B \leftarrow min\{len(Set_{k}), len(Set_{j})\}$ \\
				$overlap \leftarrow len(Set_{k} \& Set_{j}) / B $ \\
				\If{$overlap > \delta$}
				{
					$\mathbf{M}_{occ}[k-1, j-1] \leftarrow 1$ \\
					$\mathbf{M}_{occ}[j-1, k-1] \leftarrow 1$ \\
				}
			}
		}
	}
	\textbf{return} Co-occurrence Matrix $\mathbf{M}_{occ}$\\
	\caption{Co-occurrence Extraction}
	\label{algorithm}
\end{algorithm}

Assume the total number of the nodes to be $K_{occ}$. The input for the algorithm is the sentence node $U_{k}$, corpus $C_{s}$ containing redundant stop words and hyper-parameter $\delta$. The output is an adjacent matrix $\mathbf{M}_{occ} \in \mathbb{R}^{K_{occ} \times K_{occ}} $, which reflects the co-occurrence relations between nodes. As for any two nodes, we transform them into two token sets $Set_{k}$and $Set_{j}$ separately, without order and duplicate elements (Line 3 \& Line 5 in Algorithm \ref{algorithm}). We define $len(Set)$ to be the number of tokens in a set. Further, let the token overlap ratio of two sets be the co-occurrence metric (Line 7 \& Line 8). Thus, we can determine the co-occurrence relation between $U_{k}$ and $U_{j}$ if the overlap score exceeds the threshold.

So far, the connections within the sentence nodes have been explored based on the co-occurrence relations. Thus, the syntax graph is constructed reflected by the obtained adjacent matrix $\mathbf{M}_{occ} \in \mathbb{R}^{K_{occ} \times K_{occ}}$.

\subsection{Graph Transformer}
Some previous works \cite{zhang2020graph, dwivedi2020generalization} point out the drawbacks of graph neural network, such as the issue of over-smooth \cite{li2018deeper}. Therefore, we take the novel architecture of graph transformer \cite{ying2021transformers, cai2020graph} into account. After the extraction of nodes and the construction of two graphs, we feed them into the logical-aware and syntax-aware graph transformer structures respectively. 

\subsubsection{\textbf{Logical-aware Graph Transformer}}
\begin{figure}[t]
	\large
	\centering
	\includegraphics[scale=0.5]{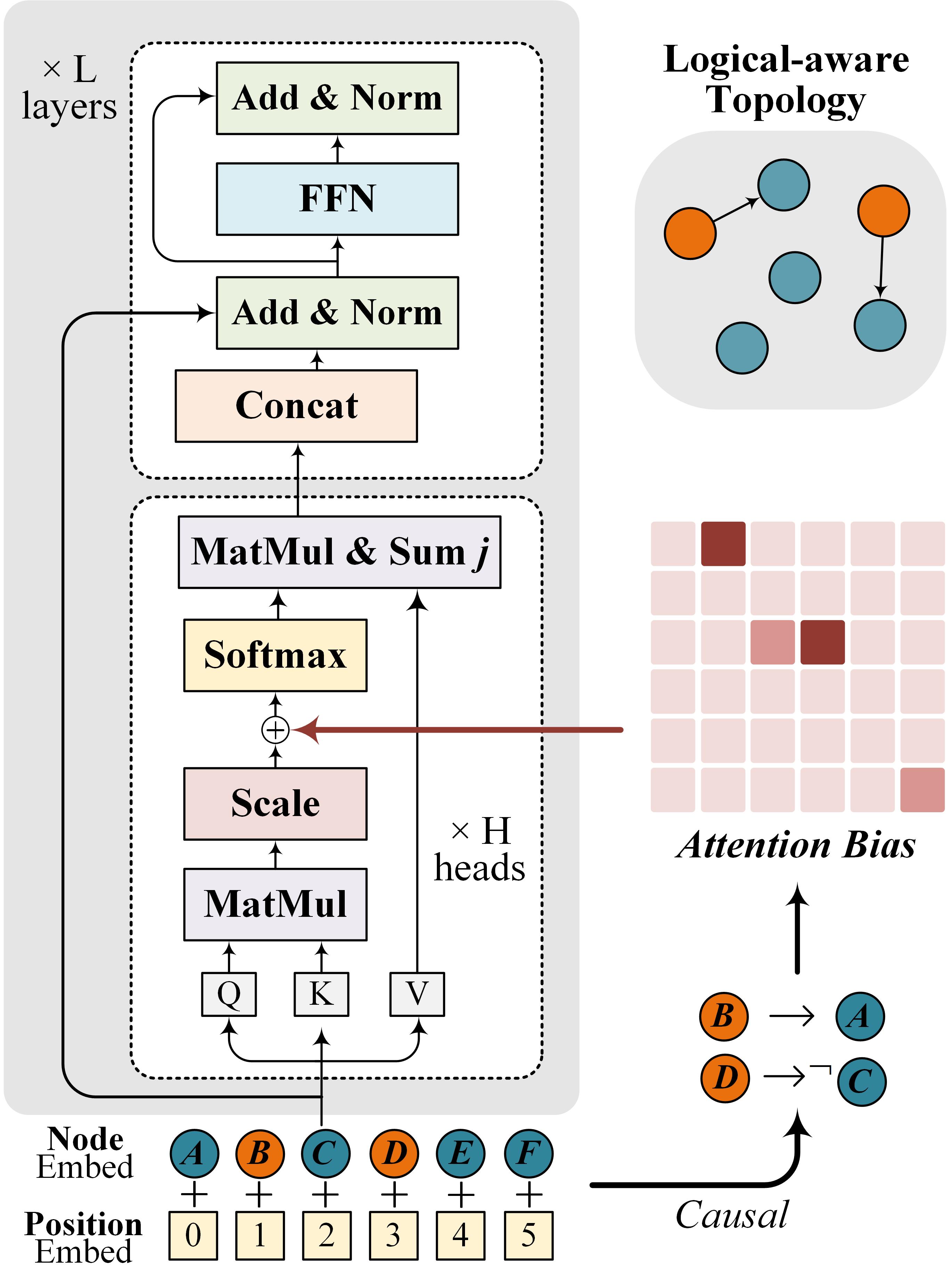}
	\vspace{-0.2cm}
	\caption{The illustration of logical-aware graph transformer. The inputs are the node sequence as well as the topology and the outputs are omitted.}
	\label{transformer}
	\vspace{-0.3cm}
\end{figure}

The simple illustration of the logical-aware graph transformer is shown in Figure \ref{transformer}. First of all, it is necessary to get the original feature embedding for each node. Given the concatenated input sequence of the $i^{th}$ question:

\begin{equation}
	{\rm Input}(c_{i},a_{i, j}) = [C L S] c_{i} [S E P] a_{i, j} [S E P],
\end{equation}
we employ the RoBERTa model \cite{liu2019roberta} as the encoder for the token-level features. For the token sequence $\{t_{1}^{(k)}, t_{2}^{(k)},..., t_{T}^{(k)}\}$ with the length $T$ of each node $U_{k}$, the obtained token embedding is represented as $\{\mathbf{v}_{t_{1}}^{(k)}, \mathbf{v}_{t_{2}}^{(k)},..., \mathbf{v}_{t_{T}}^{(k)}\}$. We take the average embedding of $T$ tokens as the original feature for node $U_{k}$:

\begin{equation}
	\mathbf{v}_{k} = \frac{1}{M}\sum\limits_{i = 1}^M {\mathbf{v}_i^{(k)}}. 
\end{equation}

To keep the original order information of nodes in the text, positional embedding is added to the node representation.

\begin{equation}
	\mathbf{V_{i}} = \mathbf{V_{o}} + PosEmbed(\mathbf{V_{o}}),
\end{equation}
where $\mathbf{V_{o}} = [\mathbf{v}_{1}; \mathbf{v}_{2}; ... ; \mathbf{v}_{K_{cas}}]$, $\mathbf{V_{o}} \in \mathbb{R}^{K_{cas}\times d}$, $d$ is the dimension of the hidden state, and $K_{cas}$ is the number of nodes. $PosEmbed(\cdot)$ provides a d-dimensional embedding for each node in the input sequence.

We feed the node representation $\mathbf{V_{i}}$ into the logical-aware graph transformer. Firstly, $\mathbf{V_{i}}$ is projected to three matrices $Q$, $K$ and $V$ of the self-attention module:

\begin{equation}
	\begin{aligned}
		Q &= \mathbf{V_{i}} \cdot{\rm  W^{Q}},\\
		K &= \mathbf{V_{i}} \cdot {\rm W^{K}},\\
		V &= \mathbf{V_{i}} \cdot {\rm W^{V}},
	\end{aligned}
\end{equation}
where ${\rm W^{Q}, W^{K}, W^{V}} \in \mathbb{R}^{d \times d_{k}}$ are projection matrices, and the obtained matrices $Q, K, V \in \mathbb{R}^{K_{cas} \times d_{k}}$. Then, we compute the attention based on the query, key and value matrices.

\begin{equation}
	\begin{aligned}
		A &= \frac{Q K^{\rm T}}{\sqrt{d_{k}}}, \\
		Att(Q, K, V) &= {\rm softmax}(A) \cdot V,
	\end{aligned}
\end{equation}
where $A \in \mathbb{R}^{K_{cas}\times K_{cas}}$ is a weight matrix for node 
pairs. From the equations, the transformer structure provides a fully connected 
setting to all nodes, which ignores the inner causal relations. Therefore, 
Logiformer employs the obtained topology information $\mathbf{M}_{cas} \in 
\mathbb{R}^{K_{cas}\times K_{cas}}$ of the logical graph as an attention bias. 
The representation of the weight matrix $A$ is adjusted as follows:

\begin{equation}
	{A}^{'} = \frac{Q K^{\rm T}}{\sqrt{d_{k}}} + \mathbf{M}_{cas}.
\end{equation}

To improve the robustness and capability of the attention module, we apply the multi-head attention mechanism with the head number $H$:

\begin{equation}
	Att_{MH}(Q, K, V) = [Head_{1}; ...; Head_{H}] \cdot {\rm W^{H}},
\end{equation}
where ${\rm W^{H}}\in \mathbb{R}^{(H*d_{k}) \times d_{k}}$ is the linear projection matrix, $Head_{i}=Att_{i}(Q, K, V)$, the input query, key and value matrices are obtained by the linear projections of ${\rm W_{i}^{Q}, W_{i}^{K}, W_{i}^{V}} \in \mathbb{R}^{d \times d_{k}}$ respectively. For simplicity, we assume $d = d_{k}$ and omit the bias term of the linear projection.

Repeating the multi-head attention for $L$ layers, we take out the hidden states of the last two layers. To enhance the robustness of the model, we make a fusion of them as the updated node features:

\begin{equation}
	\mathbf{V}_{cas} = \mathbf{V}_{cas}^{(L-1)} + \mathbf{V}_{cas}^{(L)},
\end{equation}
where $\mathbf{V}_{cas} \in \mathbb{R}^{K_{cas}\times d}$, and $\mathbf{V}_{cas}^{(L-1)}, \mathbf{V}_{cas}^{(L)} \in \mathbb{R}^{K_{cas}\times d}$ represents the hidden states of the last two layers respectively. Note that there are lots of ways of feature fusion, we only present the simple addition for illustration. 

\subsubsection{\textbf{Syntax-aware Graph Transformer}}
The general idea of the syntax-aware graph transformer is similar to that of the logical-aware graph transformer. The initialization of node features are also obtained by averaging the embedding of each token. And the positional information is kept to provide the perception of original text orders. After obtaining the weight matrix from the self-attention module, we apply the adjacent matrix $\mathbf{M}_{occ}$ to provide the attention bias for the weights. The final feature of syntax-branch node sequence is also obtained through the fusion of last two layers, represented as $\mathbf{V}_{occ} \in \mathbb{R}^{K_{occ} \times d}$.

\subsection{Decoder}
So far, we have obtained the token-level representation $\mathbf{V}_{t} \in \mathbb{R}^{N \times d}$, syntax branch node representation $\mathbf{V}_{occ} \in \mathbb{R}^{K_{occ} \times d}$ and logical branch node representation $\mathbf{V}_{cas} \in \mathbb{R}^{K_{cas} \times d}$. To make comprehensive use of the advantages of the three features, it is required to ensure dimension consistency. Therefore, we broadcast the feature of each node to all the tokens it contains. The updated features are transformed to $\mathbf{V}_{t}$, $\mathbf{V}_{occ}^{'}$, $\mathbf{V}_{cas}^{'} \in \mathbb{R}^{N \times d}$.

The above three features with the same dimension do not necessarily make equal contributions to the final prediction. It will be beneficial to automatically assign the importance. Thus, Logiformer will dynamically generate the fusion weights for $\mathbf{V}_{occ}^{'}$ and $\mathbf{V}_{cas}^{'}$. The gate parameter $\lambda$ is expressed as follows:

\begin{equation}
	\lambda = {\rm softmax}([\mathbf{V}_{occ}^{'};\mathbf{V}_{cas}^{'}]{\rm W}_{g}+b_{g}),
\end{equation}
where ${\rm W}_{g} \in \mathbb{R}^{2d \times 1}$ and $b_{g} \in \mathbb{R}^{N\times 1}$ are the weight term and bias term for the linear projection. $\mathbf{V}_{occ}^{'}$ and $\mathbf{V}_{cas}^{'}$ are concatenated at the last dimension. The obtained gate parameter $\lambda \in \mathbb{R}^{N\times 1}$ is a vector. That is to say, each token will be provided with one specific weight.

The final feature $\mathbf{V}$ can be represented in the following expression:
\begin{equation}
	\mathbf{V} = LN(\mathbf{V}_{t} + \lambda \cdot \mathbf{V}_{occ}^{'} + (1-\lambda) \cdot \mathbf{V}_{cas}^{'}),
\end{equation}
where $LN(\cdot)$ denotes the layer normalization operation. Since we do not employ the global node in the graph transformer, the global feature will not updated. To this end, Logiformer integrates the local token-level features and gets the updated global information:

\begin{equation}
	\mathbf{V}_{cls} = LN(\mathbf{V}_{t,cls} + \frac{1}{N-1}\sum\limits_{i=1}^{N-1} {(\mathbf{V}_{occ,i}^{'}+\mathbf{V}_{cas,i}^{'})}),
\end{equation}
where $\mathbf{V}_{t,cls}$ is the first token of the original token-level embedding. $\mathbf{V}_{occ,i}^{'}$ and $\mathbf{V}_{cas,i}^{'}$ represent the $i^{th}$ token embedding of the syntactic branch and logical branch feature respectively. We utilize the global feature $\mathbf{V}_{cls}$ to replace the first token feature (i.e., [cls] feature) of $\mathbf{V}$. $\mathbf{V}$ can be expressed as the concatenation of $\mathbf{V}_{cls}$, $\mathbf{V}_{context}$ and $\mathbf{V}_{option}$, that is $\mathbf{V} = [\mathbf{V}_{cls};\mathbf{V}_{context};\mathbf{V}_{option}]$.

To conduct the reasoning, the feature of the question $\mathbf{V}_{question}$ is also of great significance. Logiformer applies a simple self-attention module for the global feature $\mathbf{V}$ and question $\mathbf{V}_{question}$. The updated question embedding is expressed as:

\begin{equation}
	\mathbf{V}_{question}^{'} = {\rm softmax}(\frac{{{\mathbf{V}_{question}}{\mathbf{V}^{\rm T}}}}{{\sqrt d }})\cdot \mathbf{V}.
\end{equation}

For simplicity, the linear projections for the self-attention are omitted. At last, we concatenate the $\mathbf{V}_{cls}$, $\mathbf{V}_{context}$, $\mathbf{V}_{option}$ and $\mathbf{V}_{question}^{'}$ to get the final feature $\mathbf{V}_{final} \in \mathbb{R}^{N\times d}$:

\begin{equation}
	\mathbf{V}_{final} = [\mathbf{V}_{cls};\mathbf{V}_{context};\mathbf{V}_{option};\mathbf{V}_{question}^{'}].
\end{equation}

For each option in one example, we can get one specific final feature. They are fed into the feed forward network to obtain the scores, and we take the highest one as the predicted answer. 

\section{Experiments}
In this section, extensive experiments are conducted to compare our model with SOTA single model methods in both ReClor and LogiQA datasets. Ablation studies are followed to verify the effectiveness of the proposed modules.

\subsection{Datasets and Baselines}
\subsubsection{\textbf{Datasets}}
In this paper, we conduct the experiments on two logical reasoning datasets ReClor \cite{yu2019reclor} and LogiQA \cite{liu2020logiqa}. ReClor consists of 6,138 examples sourced from some standardized tests, while LogiQA includes totally 8,678 questions collected from National Civil Servants Examinations of China. The detailed splits of both datasets are included in Table \ref{dataset}. It can be concluded that ReClor is more diverse in the number of logical reasoning types, while LogiQA contains more examples. Both of them are challenging for the task of logical reasoning.

\begin{table}[t]
	\centering
	\caption{Detailed Splits of ReClor and LogiQA.}
	\vspace{-0.3cm}
	\begin{tabular}{p{1.7cm}|cccc}
		\toprule
		\textbf{Dataset} &\textbf{\#Train} &\textbf{\#Valid} &\textbf{\#Test} &\textbf{\#Reason Type}\\
		\hline
		ReClor & 4,638 &500 &1,000 &17\\
		LogiQA & 7,376 &651 &651 &5\\
		\bottomrule
	\end{tabular}
	\label{dataset}
	\vspace{-0.2cm}
\end{table}

\begin{table}[t]
	\caption{The tuned hyper-parameters with search scopes.}
	\vspace{-0.3cm}
	\centering
	\begin{tabular}{p{3.8cm}|cc}
		\toprule
		\textbf{Name of Parameter} & \textbf{Search Scope} & \textbf{Best}\\
		\hline
		training batchsize &\{1,2,4,8\} &2 \\
		\#epoch &\{9,10,11,12,13\} &12 \\
		\#head in graph transformer &\{4,5,6,7,8\} &5\\
		\#layer in graph transformer &\{4,5,6,7,8\} &5\\
		max sequence length &\{128,256,512\} &256 \\
		learning rate for RoBERTa &\{4e-6. 5e-6, 6e-6, 5e-5\} &5e-6 \\
		\bottomrule
	\end{tabular}
	\label{tab:hyper}
	\vspace{-0.3cm}
\end{table}

\subsubsection{\textbf{Baselines}}
To prove the superiority of our model, we mainly employ the following baselines, including the SOTA method of single model. 

\begin{itemize}
	\item \textbf{Random}: The results are based on the random predictions.
	
	\item \textbf{Human Performance}\cite{yu2019reclor, liu2020logiqa}: For ReClor, human performance is defined as the average score of different graduate students in a university on the test split. For LogiQA, the result is the average score of three post-graduate students on 500 randomly selected instances from the test split.
	
	\item \textbf{DAGN} \cite{huang2021dagn}: It proposed a discourse-aware network, which took RoBERTa-Large\cite{liu2019roberta} as the token encoder and employed GNN for the feature update.
	
	\item \textbf{FocalReasoner} \cite{ouyang2021fact}: It focused on the fact units extracted from the text and built a supergraph for the reasoning. Similar to DAGN, it also leveraged RoBERTa-Large and GNN \cite{scarselli2008graph} for the token embedding and node update respectively.
	
	\item \textbf{LReasoner} \cite{wang2021logic}: It captured the symbolic logic from the text and further extended them into natural language based on several logical equivalence laws. For the fair comparison, we take the results of the single model with RoBERTa encoder into consideration.
\end{itemize}

\subsection{Implementation Details}
All of the experiments are conducted with a single GPU of Tesla V100. For the fair comparison, the RoBERTa-large model \cite{liu2019roberta} is utilized as the encoder for text during the experiments and the hidden size is set to 1024. During the training process, the epoch number is fixed to 12 and the batchsize is set to 2 for both ReClor and LogiQA datasets. We take Adam \cite{kingma2014adam} with linearly-decayed learning rate and warm up and select peak learning rate as 5e-6. We select the model with best accuracy on the validation split to conduct the test. The details of important hyper-parameters and their search scopes are attached in Table \ref{tab:hyper}.

\subsection{Comparison Results}
Logiformer is evaluated on two logical reasoning datasets. The main results on the validation split and test split of ReClor dataset are shown in Table \ref{tab:RECLOR_main}. And the results on LogiQA dataset are shown in Table \ref{tab:LogiQA_main}. The test split of ReClor is organized into easy fold and hard fold, presented as `Test-E' and `Test-H' respectively in the table.

\begin{table}[t]
	\centering
	\caption{Experimental results on ReClor dataset. The percentage signs (\%) of accuracy values are omitted. The optimal and sub-optimal results are marked in bold and underline respectively (same for the following tables).}
	\vspace{-0.2cm}
	\begin{tabular}{p{3.2cm}|cccc}
		\toprule
		\textbf{Model} &\textbf{Valid} &\textbf{Test} &\textbf{Test-E} &\textbf{Test-H}\\
		\hline
		Random &25.00 &25.00 &25.00 &25.00 \\
		Human Performance\cite{yu2019reclor}  &- &63.00 &57.10 &67.20 \\
		BERT-Large \cite{yu2019reclor} &53.80 &49.80 &72.00 &32.30 \\
		XLNet-Large\cite{yu2019reclor} &62.00 &56.00 &75.70 &40.50 \\
		RoBERTa-Large \cite{yu2019reclor}  &62.60 & 55.60 &75.50 &40.00 \\
		DAGN \cite{huang2021dagn} &65.80 & 58.30 &75.91 &44.46 \\
		FocalReasoner \cite{ouyang2021fact} & \underline{66.80} &58.90 &77.05 &44.64 \\
		LReasoner \cite{wang2021logic} & 66.20 & \underline{62.40} &- &-\\
		\hline
		Logiformer & \textbf{68.40} & \textbf{63.50} &79.09 &51.25\\
		\bottomrule
	\end{tabular}
	\label{tab:RECLOR_main}
	\vspace{-0.1cm}
\end{table}

\begin{table}[t]
	\centering
	\caption{Experimental results on LogiQA dataset.}
	\vspace{-0.3cm}
	\begin{tabular}{p{3.5cm}|cc}
		\toprule
		\textbf{Model} &\textbf{Valid} &\textbf{Test}\\
		\hline
		Random &25.00 &25.00 \\
		Human Performance\cite{liu2020logiqa} &- &86.00 \\
		BERT-Large \cite{liu2020logiqa} &34.10 &31.03 \\
		RoBERTa-Large \cite{liu2020logiqa} &35.02 &35.33 \\
		DAGN \cite{huang2021dagn} &36.87 &39.32 \\
		FocalReasoner \cite{ouyang2021fact} &\underline{41.01} &\underline{40.25} \\
		\hline
		Logiformer &\textbf{42.24} &\textbf{42.55} \\
		\bottomrule
	\end{tabular}
	\label{tab:LogiQA_main}
	\vspace{-0.3cm}
\end{table}

\begin{table*}[t]
	\centering
	\caption{Ablation Studies. The improvements on the accuracy are marked in red.}
	\vspace{-0.2cm}
	\begin{tabular}{p{4.8cm}|cccccc|cccc}
		\toprule
		\multirow{2}*{\textbf{Model}} &\multicolumn{6}{c|}{\textbf{ReClor}} &\multicolumn{4}{c}{\textbf{LogiQA}} \\
		~ &\textbf{Valid} &$\Delta$ &\textbf{Test} &$\Delta$ &\textbf{Test-E} &\textbf{Test-H} &\textbf{Valid} &$\Delta$ &\textbf{Test} &$\Delta$\\
		\hline
		Logiformer &68.40 &- &63.50 &- &79.09 &51.25 &42.24 &- &42.55 &- \\
		\textbf{a) Graph Construction} & & & & & & & & & &\\
		\quad\quad w/o syntax graph &66.40 &\cellcolor{red!15}-2.00 &61.20 &\cellcolor{red!15}-2.30 &77.50 &48.39 &38.56 &\cellcolor{red!40}-3.68 &38.71 &\cellcolor{red!40}-3.84 \\
		\quad\quad w/o logical graph &63.60 &\cellcolor{red!40}-4.80 &59.90 &\cellcolor{red!40}-3.60 &75.00 &48.04 &38.25 &\cellcolor{red!40}-3.99 &37.63 &\cellcolor{red!40}-4.92 \\
		\textbf{b) Graph Transformer} & & & & & & & & & &\\
		\quad\quad w/o co-occurrence bias &66.80 &\cellcolor{red!15}-1.60 &62.80 &\cellcolor{red!15}-0.70 &77.05 &51.61 &41.94 &\cellcolor{red!15}-0.30 &42.55 &- \\
		\quad\quad w/o causal bias &65.20 &\cellcolor{red!40}-3.20 &63.30 
		&\cellcolor{red!15}-0.20 &76.82 &52.68 &39.94 &\cellcolor{red!15}-2.30 
		&41.47 &\cellcolor{red!15}-1.08 \\
		\quad\quad w/o both of attention biases &66.20 &\cellcolor{red!15}-2.20 &61.60 &\cellcolor{red!15}-1.90 &75.23 &50.89 &41.63 &\cellcolor{red!15}-0.61 &39.94 &\cellcolor{red!15}-2.61 \\
		\textbf{c) Decoder} & & & & & & & & & &\\
		\quad\quad w/o dynamic gates &67.00 &\cellcolor{red!15}-1.40 &61.90 &\cellcolor{red!15}-1.60 &76.14 &50.71 &41.32 &\cellcolor{red!15}-0.92 &42.55 &- \\
		\quad\quad w/o question-aware attention &66.60 &\cellcolor{red!15}-1.80 &60.40 &\cellcolor{red!40}-3.10 &76.36 &47.86 &41.63 &\cellcolor{red!15}-0.61 &42.09 &\cellcolor{red!15}-0.46 \\
		\bottomrule
	\end{tabular}
	\label{ablation}
\end{table*}

\begin{table*}[t]
	\centering
	\caption{The details of ReClor Test Split on different question types. \textbf{NA}: Necessary Assumption, \textbf{S}:Strengthen, \textbf{W}:Weaken, \textbf{I}:Implication, \textbf{CMP}:Conclusion/Main Point, \textbf{MSS}:Most Strongly Supported, \textbf{ER}:Explain or Resolve, \textbf{P}:Principle, \textbf{D}:Dispute, \textbf{R}:Role, \textbf{IF}:Identify a Flaw, \textbf{O}:Others.}
	\vspace{-0.1cm}
	\begin{tabular}{p{3.2cm}|cccccccccccc}
		\toprule
		\textbf{Model} &\textbf{NA} &\textbf{S} &\textbf{W}  &\textbf{I} &\textbf{CMP} &\textbf{MSS} &\textbf{ER} &\textbf{P} &\textbf{D} &\textbf{R} &\textbf{IF} &\textbf{O}\\
		\hline
		Logiformer &74.56 &64.89 &55.75 &45.65 &75.00 &66.07 &61.90 &69.23 &70.00 &75.00 &58.12 &60.27 \\
		\hline
		\quad\quad w/o syntax graph &70.18 &59.57 &55.75 &45.65 &66.67 &57.14 &67.86 &56.92 &56.67 &50.00 &62.39 &57.53 \\
		\quad\quad\quad\quad\quad $\Delta$ & \cellcolor{red!15}-4.38 &\cellcolor{red!15}-5.32 &- &- &\cellcolor{red!15}-8.33 &\cellcolor{red!15}-8.93 &\cellcolor{green!15}+5.96 &\cellcolor{red!15}-12.31 &\cellcolor{red!15}-13.33 &\cellcolor{red!15}-25.00 &\cellcolor{green!15}+4.27 &\cellcolor{red!15}-2.74\\
		\hline
		\quad\quad w/o logical graph &68.42 &61.70 &51.33 &41.30 &66.67 &51.79 &59.52 &55.38 &43.33 &59.38 &63.25 &65.75 \\
		\quad\quad\quad\quad\quad $\Delta$ &\cellcolor{red!15}-6.14 &\cellcolor{red!15}-3.19 &\cellcolor{red!15}-4.42 &\cellcolor{red!15}-4.34 &\cellcolor{red!15}-8.33 &\cellcolor{red!15}-14.28 &\cellcolor{red!15}-2.38 &\cellcolor{red!15}-13.85 &\cellcolor{red!15}-26.67 &\cellcolor{red!15}-15.62 &\cellcolor{green!15}+5.13 &\cellcolor{green!15}+5.48 \\
		\bottomrule
	\end{tabular}
	\label{RECLOR_test}
	\vspace{-0.3cm}
\end{table*}

For the fair comparison, we consider the results with single model and with the encoder of RoBERTa for all the baselines. Compared with the SOTA results on two logical reasoning benchmarks, our proposed model Logiformer shows excellent improvements.

On the ReClor dataset, we witness the improvements of 2.20\% and 1.10\% on the validation and test split over previous SOTA model LReasoner. Since LReasoner does not make the results on Test-E and Test-H splits public, we omit the comparison. Compared with FocalReasoner on the validation split, Logiformer shows strong generalization capability with 1.6\% and 4.6\% improvements on the validation and test split. Especially on the Test-H split, 6.61\% improvement proves our superiority for the more difficult logical reasoning questions. The most important observation is that Logiformer is the first single model with RoBERTa encoder to beat the human performance by 0.50\% on the ReClor dataset. Although the machine still falls behind humans on more challenging questions, our proposed method is positively narrowing the gaps.

On the LogiQA dataset, Logiformer outperforms the previous SOTA model Logiformer by 1.13\% and 2.30\% on the validation and test split respectively. It proves the excellent generalization capability of Logiformer. However, we also discover the huge gap between humans and the machine. In view that the context in LogiQA dataset is organized in a more structural form, humans are easier to capture the inner logic. The deep learning based models  are good at capturing the semantic changes and lack the perception of fixed logic.

\vspace{-0.3cm}
\subsection{Ablation Studies}
Considering that the architecture of Logiformer is mainly divided into three parts: a) graph construction, b) graph transformer and c) decoder, the ablation studies are also laid out from these three aspects. The experimental results are shown in Table \ref{ablation}.

Firstly, in the part of graph construction, we build syntax graph and logical 
graph based on the different node extraction strategies. We ablate the effects 
of the two graphs in turn. That is to say, we only consider one of the branches 
each time. From the results, the logical graph contributes more to the 
performance on the ReClor dataset, which improves 4.80\% and 3.60\% on the 
validation and test split respectively. The syntax graph also shows 2.00\% and 
2.30\% improvements on ReClor. As is mentioned above, we are the first to model 
the causal relations within the context in the task of logical reasoning. The 
effectiveness of logical graph also verifies our proposed method.

Secondly, we explore the impact of two attention biases on the model 
performance. Thus, we ablate the effects of one or both of attention bias 
matrices. From the results, co-occurrence bias and causal bias have different 
effects on the two splits of ReClor dataset, where the former one contributes 
more to the test split and the latter one is more helpful to the validation 
split. Meanwhile, positive effects are witnessed by applying both of the 
attention biases to the graph transformer, leading to 1.90\% and 2.61\% on the 
test split of ReClor and LogiQA respectively. Combining the ablation results 
for graph construction module, the fully connected structure of the logical 
units itself also has a positive role in the model performance. 

Thirdly, we focus on the effectiveness of two important parts in the decoder. For the proposed dynamic gate mechanism, we set each element of the gate parameter vector $\lambda \in \mathbb{R}^{N \times 1}$ to 0.5 to ablate the effect of gates. The results show that dynamic gate mechanism contributes 1.6\% improvement to the test split of ReClor, but does not have effects on that of LogiQA. It may result from the characteristics of LogiQA dataset, which require the equal contribution of syntax and logical information. For the question-aware attention, we remove the self-attention module and use the original token-level representation of the question to form the final vector. The ablation results illustrate that the update of the question feature contributes a lot to the model performance, especially for the ReClor dataset. Considering that the question types are various on the ReClor dataset, the awareness of the question sentence is of great help.

Additionally, we present the detailed results of ReClor test split on different question types and also list the corresponding ablation results of two graphs in Figure \ref{RECLOR_test}. The majority of the types witness the significant improvements, especially for \textit{Principle}, \textit{Dispute} and \textit{Role}. It illustrates that Logiformer has the advantages of inferring the hidden fact or truth within the context. A few types, such as \textit{Explain or Resolve} and \textit{Identify a Flaw}, show a downward trend. We blame this issue to the lack of modeling on negation. For example, the type of \textit{Identify a Flaw} requires the model to figure out the most weakness one from the options, which is sentimentally opposite to the most of the types. The feature distribution obtained from the current language model is insufficient to clearly distinguish the implicit opposite semantics. Therefore, the modeling of sentimentally negative questions is worth exploring in the future work.
\begin{figure}[t]
	\large
	\centering
	\includegraphics[scale=0.38]{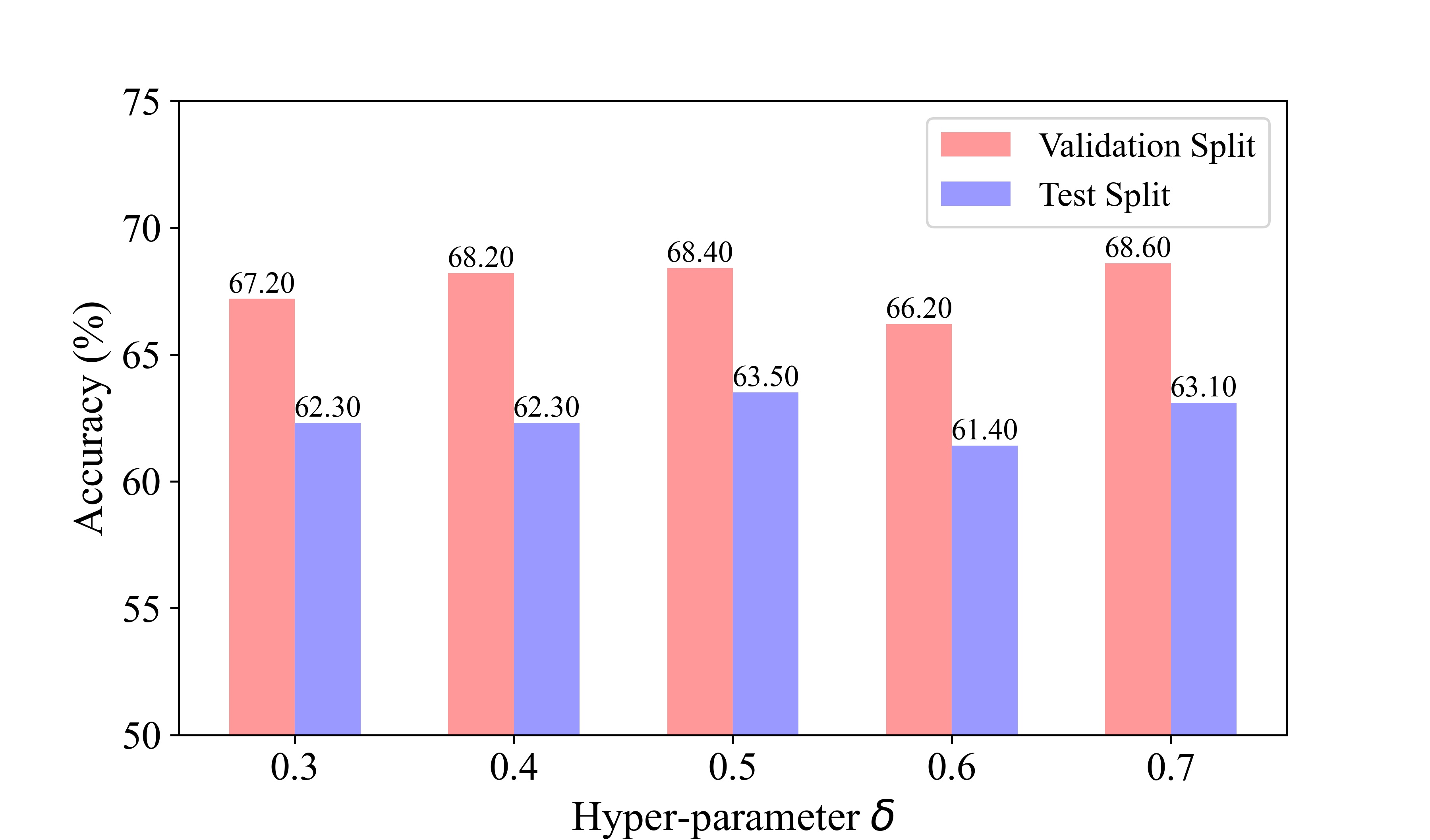}
	\vspace{-0.2cm}
	\caption{The model performances on the ReClor dataset under different $\delta$.}
	\label{delta}
	\vspace{-0.5cm}
\end{figure}

\begin{figure}[t]
	\large
	\centering
	\includegraphics[scale=0.40]{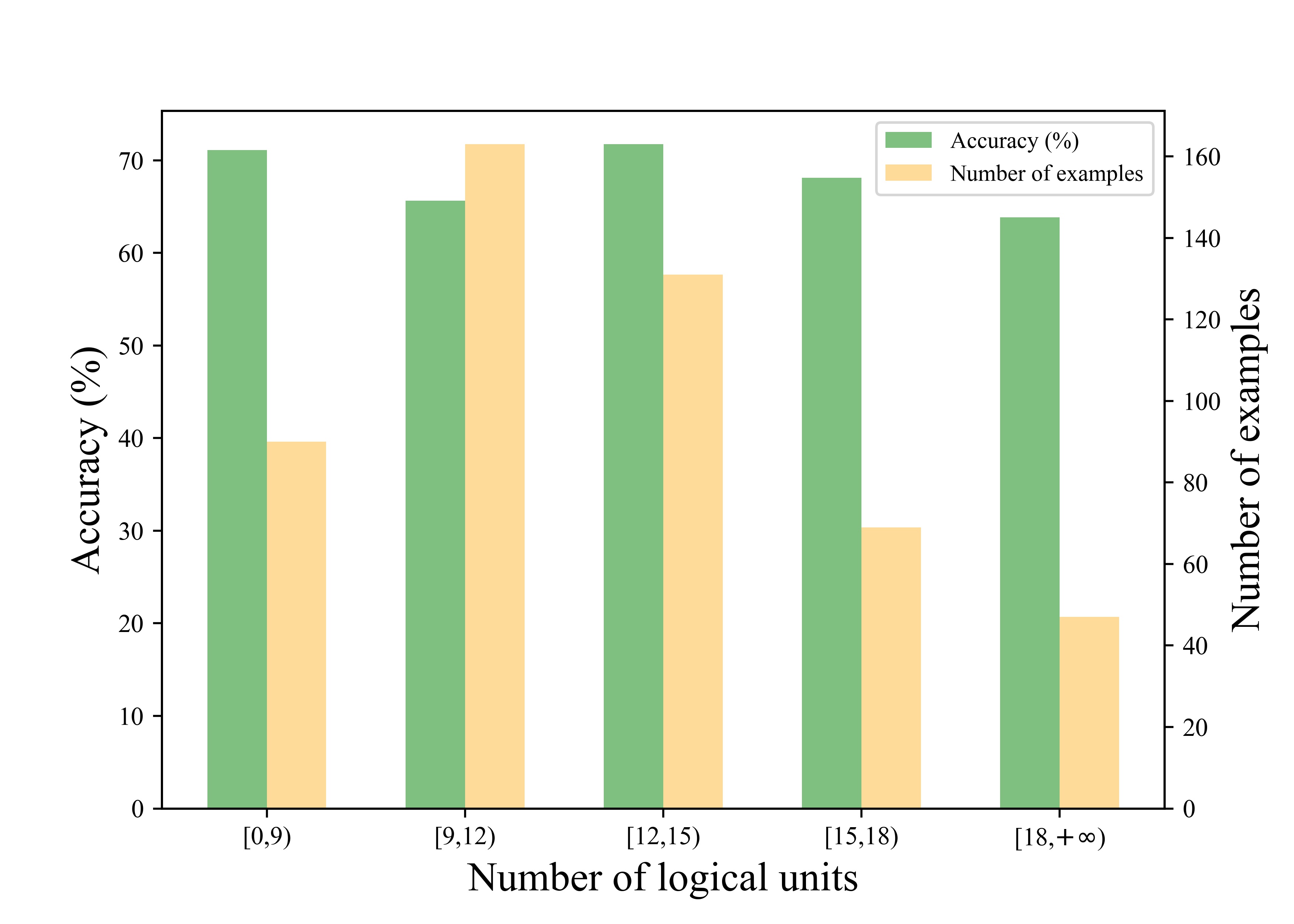}
	\vspace{-0.3cm}
	\caption{The model performances on under different numbers of logical units.}
	\label{nodenum}
	\vspace{-0.5cm}
\end{figure}

\begin{figure*}[t]
	\large
	\centering
	\includegraphics[scale=0.5]{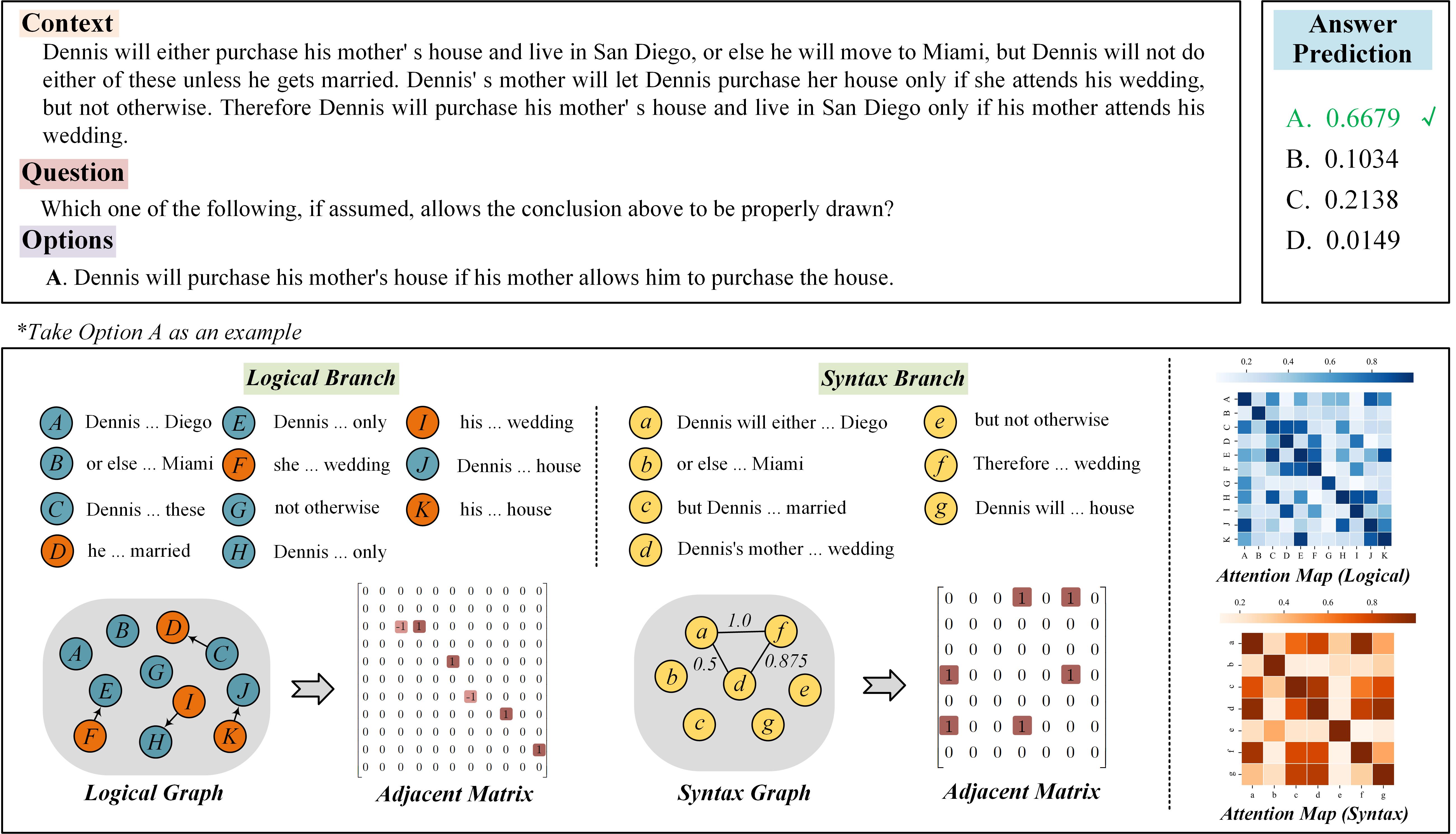}
	\caption{The illustration of an successful case. The interpretability of Logiformer lies in the logical units in text with explicit relations and the visualization of the weighted attention maps.}
	\label{casestudy}
\end{figure*}

\vspace{-0.3cm}
\subsection{Supplementary Analysis}
During the experiments, hyper-parameters are utilized in many places. Limited by the space, we only select one of them to conduct the analysis, which is the overlap threshold $\delta$ for the extraction of co-occurrence. The results of different values of $\delta$ are shown in Figure \ref{delta}.

Results illustrate that the best performance is achieved when $\delta$ is equal to 0.5. When the hyper-parameter $\delta$ drops, it means that more co-occurrence pairs will be extracted, leading to much extra noise. When $\delta$ reaches 0.7, the number of co-occurrence relations is limited, the performance of the model still maintains at a high level. It further proves the robustness of Logiformer.

Also in Logiformer, logical units are important parts. Thus, we will study the influence of the number of logical units on the accuracy. We conduct the analysis on the validation split of ReClor dataset. The number of logical units are obtained based on the split of relation connectives and punctuations. The results are presented in Figure \ref{nodenum}. The green bar represents the accuracy and the yellow bar is the number of examples. From the statistics, the examples with 9-12 logical units account for the largest proportion. Most importantly, the accuracy remains stable for the different number of logical units. On one hand, it proves the effectiveness of the split of logical units. On another hand, we attribute the results to the employment of the graph transformer. Fully connected structure tackles the issue of long distance dependency, which reduces the impact of the increase in the number of logical units.

\vspace{-0.1cm}
\section{Case Study}
In Figure \ref{casestudy}, we present a successful case on the validation split 
of the ReClor dataset to illustrate the logical reasoning process in 
Logiformer. The basis of the reasoning process is the two constructed graphs 
from the logical branch and the syntax branch. In this case, Logiformer 
extracts 11 logical units (named from $A$ to $K$) based on punctuations and 
relation connectives for the logical branch. The split results are consistent 
with our expectation, and among them 4 pairs of causal units ($C-D$, $E-F$, 
$H-I$, $J-K$) are detected. For the syntax branch, the concatenated text is 
split into 7 sentence nodes (named from $a$ to $g$). Among them, 3 logical 
units ($a,d,f$) are detected as the co-occurrence relations. The topology of 
the two graphs provides the explicit understanding of the text, which is a key 
point to the interpretability of Logiformer. In addition, we present the 
attention maps in the final layer of the graph transformers from both branches 
(blue one for logical branch, orange one for syntax branch). The data in the 
attention matrices is mapped to the range of [0,1] for better illustration. 
Darker color indicates the stronger correlations between two logical units. The 
weighted attention maps well reflect the relations and provide a boarder view 
for interpretability.

Meanwhile, we observe a drawback in this case. \textit{he gets married} and \textit{his wedding} are two similar expressions in semantic, indicating the similar meanings. However, they have no overlap of words and are not detected as the co-occurrence relation. This detail is worthy of studying in the future, which is beneficial to the fine-grained understanding of the logical text.

\vspace{-0.1cm}
\section{Conclusion and Future Work}
We propose a two-branch graph transformer network for logical reasoning of 
text, which is named as Logiformer. Firstly, we introduce two different 
strategies to construct the logical graph and syntax graph respectively. 
Especially for the logical graph, we are the first to model both causal 
relations and negations in the logical reasoning task. Secondly, we feed the 
extracted node sequences to the fully connected graph transformer for each 
graph. The topology of the graph is utilized to form the attention bias for the 
self-attention layers. Thirdly, a dynamic gate mechanism is applied to make a 
fusion of the features from two branches. To improve the awareness of different 
question types, the question feature is updated based on the self-attention 
module. Finally, the concatenated text sequence is passed through the feed 
forward layer and obtains the answer prediction. The whole reasoning process 
provides the interpretability, reflected by logical units with explicit 
relations and the visualization of the attention maps.

In the future, we will explore the role of question to further improve the interpretability \cite{thayaparan2020survey}. Also, we are interested in extending the logical expressions based on contrastive learning, like \cite{lin2021contrastive}.

\begin{acks}
	This work was supported by National Key Research and Development Program of 
	China (2020AAA0108800),
	National Natural Science Foundation of China (62137002, 61937001, 62176209, 
	62176207, 61877050, 62106190, 62192781 and 62050194),
	Innovative Research Group of the National Natural Science Foundation of 
	China (61721002),
	Innovation Research Team of Ministry of Education (IRT\_17R86),
	The National Social Science Fund of China(18XXW005),
	Consulting research project of Chinese academy of engineering ``The Online 
	and Offline Mixed Educational Service System for `The Belt and Road' 
	Training in MOOC China'',
	China Postdoctoral Science Foundation (2020M683493),
	Project of China Knowledge Centre for Engineering Science and Technology,
    ``LENOVO-XJTU'' Intelligent Industry Joint Laboratory Project,
	and the Fundamental Research Funds for the Central Universities 
	(xzy022021048, xpt012021005, xhj032021013-02).
\end{acks}

\newpage
\bibliographystyle{ACM-Reference-Format}
\bibliography{reference}


\begin{thebibliography}{41}


\ifx \showCODEN    \undefined \def \showCODEN     #1{\unskip}     \fi
\ifx \showDOI      \undefined \def \showDOI       #1{#1}\fi
\ifx \showISBNx    \undefined \def \showISBNx     #1{\unskip}     \fi
\ifx \showISBNxiii \undefined \def \showISBNxiii  #1{\unskip}     \fi
\ifx \showISSN     \undefined \def \showISSN      #1{\unskip}     \fi
\ifx \showLCCN     \undefined \def \showLCCN      #1{\unskip}     \fi
\ifx \shownote     \undefined \def \shownote      #1{#1}          \fi
\ifx \showarticletitle \undefined \def \showarticletitle #1{#1}   \fi
\ifx \showURL      \undefined \def \showURL       {\relax}        \fi
\providecommand\bibfield[2]{#2}
\providecommand\bibinfo[2]{#2}
\providecommand\natexlab[1]{#1}
\providecommand\showeprint[2][]{arXiv:#2}

\bibitem[\protect\citeauthoryear{Bowman, Angeli, Potts, and Manning}{Bowman
  et~al\mbox{.}}{2015}]%
        {bowman2015large}
\bibfield{author}{\bibinfo{person}{Samuel~R Bowman}, \bibinfo{person}{Gabor
  Angeli}, \bibinfo{person}{Christopher Potts}, {and}
  \bibinfo{person}{Christopher~D Manning}.} \bibinfo{year}{2015}\natexlab{}.
\newblock \showarticletitle{A large annotated corpus for learning natural
  language inference}.
\newblock \bibinfo{journal}{\emph{arXiv preprint arXiv:1508.05326}}
  (\bibinfo{year}{2015}).
\newblock


\bibitem[\protect\citeauthoryear{Brown, Mann, Ryder, Subbiah, Kaplan, Dhariwal,
  Neelakantan, Shyam, Sastry, Askell, et~al\mbox{.}}{Brown
  et~al\mbox{.}}{2020}]%
        {brown2020language}
\bibfield{author}{\bibinfo{person}{Tom~B Brown}, \bibinfo{person}{Benjamin
  Mann}, \bibinfo{person}{Nick Ryder}, \bibinfo{person}{Melanie Subbiah},
  \bibinfo{person}{Jared Kaplan}, \bibinfo{person}{Prafulla Dhariwal},
  \bibinfo{person}{Arvind Neelakantan}, \bibinfo{person}{Pranav Shyam},
  \bibinfo{person}{Girish Sastry}, \bibinfo{person}{Amanda Askell},
  {et~al\mbox{.}}} \bibinfo{year}{2020}\natexlab{}.
\newblock \showarticletitle{Language models are few-shot learners}.
\newblock \bibinfo{journal}{\emph{arXiv preprint arXiv:2005.14165}}
  (\bibinfo{year}{2020}).
\newblock


\bibitem[\protect\citeauthoryear{Cai and Lam}{Cai and Lam}{2020}]%
        {cai2020graph}
\bibfield{author}{\bibinfo{person}{Deng Cai} {and} \bibinfo{person}{Wai Lam}.}
  \bibinfo{year}{2020}\natexlab{}.
\newblock \showarticletitle{Graph transformer for graph-to-sequence learning}.
  In \bibinfo{booktitle}{\emph{Proceedings of the AAAI Conference on Artificial
  Intelligence}}, Vol.~\bibinfo{volume}{34}. \bibinfo{pages}{7464--7471}.
\newblock


\bibitem[\protect\citeauthoryear{Chowdhury}{Chowdhury}{2003}]%
        {chowdhury2003natural}
\bibfield{author}{\bibinfo{person}{Gobinda~G Chowdhury}.}
  \bibinfo{year}{2003}\natexlab{}.
\newblock \showarticletitle{Natural language processing}.
\newblock \bibinfo{journal}{\emph{Annual review of information science and
  technology}} \bibinfo{volume}{37}, \bibinfo{number}{1}
  (\bibinfo{year}{2003}), \bibinfo{pages}{51--89}.
\newblock


\bibitem[\protect\citeauthoryear{Clark, Luong, Le, and Manning}{Clark
  et~al\mbox{.}}{2020}]%
        {clark2020electra}
\bibfield{author}{\bibinfo{person}{Kevin Clark}, \bibinfo{person}{Minh-Thang
  Luong}, \bibinfo{person}{Quoc~V Le}, {and} \bibinfo{person}{Christopher~D
  Manning}.} \bibinfo{year}{2020}\natexlab{}.
\newblock \showarticletitle{Electra: Pre-training text encoders as
  discriminators rather than generators}.
\newblock \bibinfo{journal}{\emph{arXiv preprint arXiv:2003.10555}}
  (\bibinfo{year}{2020}).
\newblock


\bibitem[\protect\citeauthoryear{Devlin, Chang, Lee, and Toutanova}{Devlin
  et~al\mbox{.}}{2018}]%
        {devlin2018bert}
\bibfield{author}{\bibinfo{person}{Jacob Devlin}, \bibinfo{person}{Ming-Wei
  Chang}, \bibinfo{person}{Kenton Lee}, {and} \bibinfo{person}{Kristina
  Toutanova}.} \bibinfo{year}{2018}\natexlab{}.
\newblock \showarticletitle{Bert: Pre-training of deep bidirectional
  transformers for language understanding}.
\newblock \bibinfo{journal}{\emph{arXiv preprint arXiv:1810.04805}}
  (\bibinfo{year}{2018}).
\newblock


\bibitem[\protect\citeauthoryear{Dhingra, Liu, Yang, Cohen, and
  Salakhutdinov}{Dhingra et~al\mbox{.}}{2016}]%
        {dhingra2016gated}
\bibfield{author}{\bibinfo{person}{Bhuwan Dhingra}, \bibinfo{person}{Hanxiao
  Liu}, \bibinfo{person}{Zhilin Yang}, \bibinfo{person}{William~W Cohen}, {and}
  \bibinfo{person}{Ruslan Salakhutdinov}.} \bibinfo{year}{2016}\natexlab{}.
\newblock \showarticletitle{Gated-attention readers for text comprehension}.
\newblock \bibinfo{journal}{\emph{arXiv preprint arXiv:1606.01549}}
  (\bibinfo{year}{2016}).
\newblock


\bibitem[\protect\citeauthoryear{Dwivedi and Bresson}{Dwivedi and
  Bresson}{2020}]%
        {dwivedi2020generalization}
\bibfield{author}{\bibinfo{person}{Vijay~Prakash Dwivedi} {and}
  \bibinfo{person}{Xavier Bresson}.} \bibinfo{year}{2020}\natexlab{}.
\newblock \showarticletitle{A generalization of transformer networks to
  graphs}.
\newblock \bibinfo{journal}{\emph{arXiv preprint arXiv:2012.09699}}
  (\bibinfo{year}{2020}).
\newblock


\bibitem[\protect\citeauthoryear{Hirschberg and Manning}{Hirschberg and
  Manning}{2015}]%
        {hirschberg2015advances}
\bibfield{author}{\bibinfo{person}{Julia Hirschberg} {and}
  \bibinfo{person}{Christopher~D Manning}.} \bibinfo{year}{2015}\natexlab{}.
\newblock \showarticletitle{Advances in natural language processing}.
\newblock \bibinfo{journal}{\emph{Science}} \bibinfo{volume}{349},
  \bibinfo{number}{6245} (\bibinfo{year}{2015}), \bibinfo{pages}{261--266}.
\newblock


\bibitem[\protect\citeauthoryear{Hirschman and Gaizauskas}{Hirschman and
  Gaizauskas}{2001}]%
        {hirschman2001natural}
\bibfield{author}{\bibinfo{person}{Lynette Hirschman} {and}
  \bibinfo{person}{Robert Gaizauskas}.} \bibinfo{year}{2001}\natexlab{}.
\newblock \showarticletitle{Natural language question answering: the view from
  here}.
\newblock \bibinfo{journal}{\emph{natural language engineering}}
  \bibinfo{volume}{7}, \bibinfo{number}{4} (\bibinfo{year}{2001}),
  \bibinfo{pages}{275--300}.
\newblock


\bibitem[\protect\citeauthoryear{Hockett}{Hockett}{1947}]%
        {hockett1947problems}
\bibfield{author}{\bibinfo{person}{Charles~F Hockett}.}
  \bibinfo{year}{1947}\natexlab{}.
\newblock \showarticletitle{Problems of morphemic analysis}.
\newblock \bibinfo{journal}{\emph{Language}} \bibinfo{volume}{23},
  \bibinfo{number}{4} (\bibinfo{year}{1947}), \bibinfo{pages}{321--343}.
\newblock


\bibitem[\protect\citeauthoryear{Huang, Fang, Cao, Wang, and Liang}{Huang
  et~al\mbox{.}}{2021}]%
        {huang2021dagn}
\bibfield{author}{\bibinfo{person}{Yinya Huang}, \bibinfo{person}{Meng Fang},
  \bibinfo{person}{Yu Cao}, \bibinfo{person}{Liwei Wang}, {and}
  \bibinfo{person}{Xiaodan Liang}.} \bibinfo{year}{2021}\natexlab{}.
\newblock \showarticletitle{DAGN: Discourse-Aware Graph Network for Logical
  Reasoning}. In \bibinfo{booktitle}{\emph{Proceedings of the Conference of the
  North American Chapter of the Association for Computational Linguistics
  (NAACL)}}. \bibinfo{pages}{5848--5855}.
\newblock


\bibitem[\protect\citeauthoryear{Kingma and Ba}{Kingma and Ba}{2014}]%
        {kingma2014adam}
\bibfield{author}{\bibinfo{person}{Diederik~P Kingma} {and}
  \bibinfo{person}{Jimmy Ba}.} \bibinfo{year}{2014}\natexlab{}.
\newblock \showarticletitle{Adam: A method for stochastic optimization}.
\newblock \bibinfo{journal}{\emph{arXiv preprint arXiv:1412.6980}}
  (\bibinfo{year}{2014}).
\newblock


\bibitem[\protect\citeauthoryear{Lai, Xie, Liu, Yang, and Hovy}{Lai
  et~al\mbox{.}}{2017}]%
        {lai2017race}
\bibfield{author}{\bibinfo{person}{Guokun Lai}, \bibinfo{person}{Qizhe Xie},
  \bibinfo{person}{Hanxiao Liu}, \bibinfo{person}{Yiming Yang}, {and}
  \bibinfo{person}{Eduard Hovy}.} \bibinfo{year}{2017}\natexlab{}.
\newblock \showarticletitle{Race: Large-scale reading comprehension dataset
  from examinations}.
\newblock \bibinfo{journal}{\emph{arXiv preprint arXiv:1704.04683}}
  (\bibinfo{year}{2017}).
\newblock


\bibitem[\protect\citeauthoryear{Li, Han, and Wu}{Li et~al\mbox{.}}{2018}]%
        {li2018deeper}
\bibfield{author}{\bibinfo{person}{Qimai Li}, \bibinfo{person}{Zhichao Han},
  {and} \bibinfo{person}{Xiao-Ming Wu}.} \bibinfo{year}{2018}\natexlab{}.
\newblock \showarticletitle{Deeper insights into graph convolutional networks
  for semi-supervised learning}. In \bibinfo{booktitle}{\emph{Thirty-Second
  AAAI conference on artificial intelligence}}.
\newblock


\bibitem[\protect\citeauthoryear{Lin, Liu, Zhang, Pan, Hu, Xu, and Zeng}{Lin
  et~al\mbox{.}}{2021}]%
        {lin2021contrastive}
\bibfield{author}{\bibinfo{person}{Qika Lin}, \bibinfo{person}{Jun Liu},
  \bibinfo{person}{Lingling Zhang}, \bibinfo{person}{Yudai Pan},
  \bibinfo{person}{Xin Hu}, \bibinfo{person}{Fangzhi Xu}, {and}
  \bibinfo{person}{Hongwei Zeng}.} \bibinfo{year}{2021}\natexlab{}.
\newblock \showarticletitle{Contrastive Graph Representations for Logical
  Formulas Embedding}.
\newblock \bibinfo{journal}{\emph{IEEE Transactions on Knowledge and Data
  Engineering (TKDE)}} (\bibinfo{year}{2021}).
\newblock


\bibitem[\protect\citeauthoryear{Liu, Cui, Liu, Huang, Wang, and Zhang}{Liu
  et~al\mbox{.}}{2020}]%
        {liu2020logiqa}
\bibfield{author}{\bibinfo{person}{Jian Liu}, \bibinfo{person}{Leyang Cui},
  \bibinfo{person}{Hanmeng Liu}, \bibinfo{person}{Dandan Huang},
  \bibinfo{person}{Yile Wang}, {and} \bibinfo{person}{Yue Zhang}.}
  \bibinfo{year}{2020}\natexlab{}.
\newblock \showarticletitle{Logiqa: A challenge dataset for machine reading
  comprehension with logical reasoning}.
\newblock \bibinfo{journal}{\emph{arXiv preprint arXiv:2007.08124}}
  (\bibinfo{year}{2020}).
\newblock


\bibitem[\protect\citeauthoryear{Liu, Zhang, Zhang, Wang, and Zhang}{Liu
  et~al\mbox{.}}{2019b}]%
        {liu2019neural}
\bibfield{author}{\bibinfo{person}{Shanshan Liu}, \bibinfo{person}{Xin Zhang},
  \bibinfo{person}{Sheng Zhang}, \bibinfo{person}{Hui Wang}, {and}
  \bibinfo{person}{Weiming Zhang}.} \bibinfo{year}{2019}\natexlab{b}.
\newblock \showarticletitle{Neural machine reading comprehension: Methods and
  trends}.
\newblock \bibinfo{journal}{\emph{Applied Sciences}} \bibinfo{volume}{9},
  \bibinfo{number}{18} (\bibinfo{year}{2019}), \bibinfo{pages}{3698}.
\newblock


\bibitem[\protect\citeauthoryear{Liu, Ott, Goyal, Du, Joshi, Chen, Levy, Lewis,
  Zettlemoyer, and Stoyanov}{Liu et~al\mbox{.}}{2019a}]%
        {liu2019roberta}
\bibfield{author}{\bibinfo{person}{Yinhan Liu}, \bibinfo{person}{Myle Ott},
  \bibinfo{person}{Naman Goyal}, \bibinfo{person}{Jingfei Du},
  \bibinfo{person}{Mandar Joshi}, \bibinfo{person}{Danqi Chen},
  \bibinfo{person}{Omer Levy}, \bibinfo{person}{Mike Lewis},
  \bibinfo{person}{Luke Zettlemoyer}, {and} \bibinfo{person}{Veselin
  Stoyanov}.} \bibinfo{year}{2019}\natexlab{a}.
\newblock \showarticletitle{Roberta: A robustly optimized bert pretraining
  approach}.
\newblock \bibinfo{journal}{\emph{arXiv preprint arXiv:1907.11692}}
  (\bibinfo{year}{2019}).
\newblock


\bibitem[\protect\citeauthoryear{Mihaylov, Clark, Khot, and Sabharwal}{Mihaylov
  et~al\mbox{.}}{2018}]%
        {mihaylov2018can}
\bibfield{author}{\bibinfo{person}{Todor Mihaylov}, \bibinfo{person}{Peter
  Clark}, \bibinfo{person}{Tushar Khot}, {and} \bibinfo{person}{Ashish
  Sabharwal}.} \bibinfo{year}{2018}\natexlab{}.
\newblock \showarticletitle{Can a suit of armor conduct electricity? a new
  dataset for open book question answering}.
\newblock \bibinfo{journal}{\emph{arXiv preprint arXiv:1809.02789}}
  (\bibinfo{year}{2018}).
\newblock


\bibitem[\protect\citeauthoryear{Ouyang, Zhang, and Zhao}{Ouyang
  et~al\mbox{.}}{2021}]%
        {ouyang2021fact}
\bibfield{author}{\bibinfo{person}{Siru Ouyang}, \bibinfo{person}{Zhuosheng
  Zhang}, {and} \bibinfo{person}{Hai Zhao}.} \bibinfo{year}{2021}\natexlab{}.
\newblock \showarticletitle{Fact-driven Logical Reasoning}.
\newblock \bibinfo{journal}{\emph{arXiv preprint arXiv:2105.10334}}
  (\bibinfo{year}{2021}).
\newblock


\bibitem[\protect\citeauthoryear{Prasad, Dinesh, Lee, Miltsakaki, Robaldo,
  Joshi, and Webber}{Prasad et~al\mbox{.}}{2008}]%
        {prasad2008penn}
\bibfield{author}{\bibinfo{person}{Rashmi Prasad}, \bibinfo{person}{Nikhil
  Dinesh}, \bibinfo{person}{Alan Lee}, \bibinfo{person}{Eleni Miltsakaki},
  \bibinfo{person}{Livio Robaldo}, \bibinfo{person}{Aravind~K Joshi}, {and}
  \bibinfo{person}{Bonnie~L Webber}.} \bibinfo{year}{2008}\natexlab{}.
\newblock \showarticletitle{The Penn Discourse TreeBank 2.0.}. In
  \bibinfo{booktitle}{\emph{LREC}}. Citeseer.
\newblock


\bibitem[\protect\citeauthoryear{Rajpurkar, Jia, and Liang}{Rajpurkar
  et~al\mbox{.}}{2018}]%
        {rajpurkar2018know}
\bibfield{author}{\bibinfo{person}{Pranav Rajpurkar}, \bibinfo{person}{Robin
  Jia}, {and} \bibinfo{person}{Percy Liang}.} \bibinfo{year}{2018}\natexlab{}.
\newblock \showarticletitle{Know what you don't know: Unanswerable questions
  for SQuAD}.
\newblock \bibinfo{journal}{\emph{arXiv preprint arXiv:1806.03822}}
  (\bibinfo{year}{2018}).
\newblock


\bibitem[\protect\citeauthoryear{Rajpurkar, Zhang, Lopyrev, and
  Liang}{Rajpurkar et~al\mbox{.}}{2016}]%
        {rajpurkar2016squad}
\bibfield{author}{\bibinfo{person}{Pranav Rajpurkar}, \bibinfo{person}{Jian
  Zhang}, \bibinfo{person}{Konstantin Lopyrev}, {and} \bibinfo{person}{Percy
  Liang}.} \bibinfo{year}{2016}\natexlab{}.
\newblock \showarticletitle{Squad: 100,000+ questions for machine comprehension
  of text}.
\newblock \bibinfo{journal}{\emph{arXiv preprint arXiv:1606.05250}}
  (\bibinfo{year}{2016}).
\newblock


\bibitem[\protect\citeauthoryear{Salha, Hennequin, and Vazirgiannis}{Salha
  et~al\mbox{.}}{2020}]%
        {salha2020simple}
\bibfield{author}{\bibinfo{person}{Guillaume Salha}, \bibinfo{person}{Romain
  Hennequin}, {and} \bibinfo{person}{Michalis Vazirgiannis}.}
  \bibinfo{year}{2020}\natexlab{}.
\newblock \showarticletitle{Simple and effective graph autoencoders with
  one-hop linear models}.
\newblock \bibinfo{journal}{\emph{arXiv preprint arXiv:2001.07614}}
  (\bibinfo{year}{2020}).
\newblock


\bibitem[\protect\citeauthoryear{Scarselli, Gori, Tsoi, Hagenbuchner, and
  Monfardini}{Scarselli et~al\mbox{.}}{2008}]%
        {scarselli2008graph}
\bibfield{author}{\bibinfo{person}{Franco Scarselli}, \bibinfo{person}{Marco
  Gori}, \bibinfo{person}{Ah~Chung Tsoi}, \bibinfo{person}{Markus
  Hagenbuchner}, {and} \bibinfo{person}{Gabriele Monfardini}.}
  \bibinfo{year}{2008}\natexlab{}.
\newblock \showarticletitle{The graph neural network model}.
\newblock \bibinfo{journal}{\emph{IEEE transactions on neural networks (TNN)}}
  \bibinfo{volume}{20}, \bibinfo{number}{1} (\bibinfo{year}{2008}),
  \bibinfo{pages}{61--80}.
\newblock


\bibitem[\protect\citeauthoryear{Seo, Kembhavi, Farhadi, and Hajishirzi}{Seo
  et~al\mbox{.}}{2016}]%
        {seo2016bidirectional}
\bibfield{author}{\bibinfo{person}{Minjoon Seo}, \bibinfo{person}{Aniruddha
  Kembhavi}, \bibinfo{person}{Ali Farhadi}, {and} \bibinfo{person}{Hannaneh
  Hajishirzi}.} \bibinfo{year}{2016}\natexlab{}.
\newblock \showarticletitle{Bidirectional attention flow for machine
  comprehension}.
\newblock \bibinfo{journal}{\emph{arXiv preprint arXiv:1611.01603}}
  (\bibinfo{year}{2016}).
\newblock


\bibitem[\protect\citeauthoryear{Storks, Gao, and Chai}{Storks
  et~al\mbox{.}}{2019}]%
        {storks2019recent}
\bibfield{author}{\bibinfo{person}{Shane Storks}, \bibinfo{person}{Qiaozi Gao},
  {and} \bibinfo{person}{Joyce~Y Chai}.} \bibinfo{year}{2019}\natexlab{}.
\newblock \showarticletitle{Recent advances in natural language inference: A
  survey of benchmarks, resources, and approaches}.
\newblock \bibinfo{journal}{\emph{arXiv preprint arXiv:1904.01172}}
  (\bibinfo{year}{2019}).
\newblock


\bibitem[\protect\citeauthoryear{Thayaparan, Valentino, and Freitas}{Thayaparan
  et~al\mbox{.}}{2020}]%
        {thayaparan2020survey}
\bibfield{author}{\bibinfo{person}{Mokanarangan Thayaparan},
  \bibinfo{person}{Marco Valentino}, {and} \bibinfo{person}{Andr{\'e}
  Freitas}.} \bibinfo{year}{2020}\natexlab{}.
\newblock \showarticletitle{A survey on explainability in machine reading
  comprehension}.
\newblock \bibinfo{journal}{\emph{arXiv preprint arXiv:2010.00389}}
  (\bibinfo{year}{2020}).
\newblock


\bibitem[\protect\citeauthoryear{Vaswani, Shazeer, Parmar, Uszkoreit, Jones,
  Gomez, Kaiser, and Polosukhin}{Vaswani et~al\mbox{.}}{2017}]%
        {vaswani2017attention}
\bibfield{author}{\bibinfo{person}{Ashish Vaswani}, \bibinfo{person}{Noam
  Shazeer}, \bibinfo{person}{Niki Parmar}, \bibinfo{person}{Jakob Uszkoreit},
  \bibinfo{person}{Llion Jones}, \bibinfo{person}{Aidan~N Gomez},
  \bibinfo{person}{{\L}ukasz Kaiser}, {and} \bibinfo{person}{Illia
  Polosukhin}.} \bibinfo{year}{2017}\natexlab{}.
\newblock \showarticletitle{Attention is all you need}. In
  \bibinfo{booktitle}{\emph{Advances in neural information processing systems
  (NIPS)}}. \bibinfo{pages}{5998--6008}.
\newblock


\bibitem[\protect\citeauthoryear{Wang, Zhong, Tang, Wei, Fan, Jiang, Zhou, and
  Duan}{Wang et~al\mbox{.}}{2021}]%
        {wang2021logic}
\bibfield{author}{\bibinfo{person}{Siyuan Wang}, \bibinfo{person}{Wanjun
  Zhong}, \bibinfo{person}{Duyu Tang}, \bibinfo{person}{Zhongyu Wei},
  \bibinfo{person}{Zhihao Fan}, \bibinfo{person}{Daxin Jiang},
  \bibinfo{person}{Ming Zhou}, {and} \bibinfo{person}{Nan Duan}.}
  \bibinfo{year}{2021}\natexlab{}.
\newblock \showarticletitle{Logic-Driven Context Extension and Data
  Augmentation for Logical Reasoning of Text}.
\newblock \bibinfo{journal}{\emph{arXiv preprint arXiv:2105.03659}}
  (\bibinfo{year}{2021}).
\newblock


\bibitem[\protect\citeauthoryear{Xu, Gan, Cheng, and Liu}{Xu
  et~al\mbox{.}}{2019}]%
        {xu2019discourse}
\bibfield{author}{\bibinfo{person}{Jiacheng Xu}, \bibinfo{person}{Zhe Gan},
  \bibinfo{person}{Yu Cheng}, {and} \bibinfo{person}{Jingjing Liu}.}
  \bibinfo{year}{2019}\natexlab{}.
\newblock \showarticletitle{Discourse-aware neural extractive text
  summarization}.
\newblock \bibinfo{journal}{\emph{arXiv preprint arXiv:1910.14142}}
  (\bibinfo{year}{2019}).
\newblock


\bibitem[\protect\citeauthoryear{Yang, Dai, Yang, Carbonell, Salakhutdinov, and
  Le}{Yang et~al\mbox{.}}{2019}]%
        {yang2019xlnet}
\bibfield{author}{\bibinfo{person}{Zhilin Yang}, \bibinfo{person}{Zihang Dai},
  \bibinfo{person}{Yiming Yang}, \bibinfo{person}{Jaime Carbonell},
  \bibinfo{person}{Russ~R Salakhutdinov}, {and} \bibinfo{person}{Quoc~V Le}.}
  \bibinfo{year}{2019}\natexlab{}.
\newblock \showarticletitle{Xlnet: Generalized autoregressive pretraining for
  language understanding}.
\newblock \bibinfo{journal}{\emph{Advances in neural information processing
  systems (NIPS)}}  \bibinfo{volume}{32} (\bibinfo{year}{2019}).
\newblock


\bibitem[\protect\citeauthoryear{Yang, Qi, Zhang, Bengio, Cohen, Salakhutdinov,
  and Manning}{Yang et~al\mbox{.}}{2018}]%
        {yang2018hotpotqa}
\bibfield{author}{\bibinfo{person}{Zhilin Yang}, \bibinfo{person}{Peng Qi},
  \bibinfo{person}{Saizheng Zhang}, \bibinfo{person}{Yoshua Bengio},
  \bibinfo{person}{William~W Cohen}, \bibinfo{person}{Ruslan Salakhutdinov},
  {and} \bibinfo{person}{Christopher~D Manning}.}
  \bibinfo{year}{2018}\natexlab{}.
\newblock \showarticletitle{Hotpotqa: A dataset for diverse, explainable
  multi-hop question answering}.
\newblock \bibinfo{journal}{\emph{arXiv preprint arXiv:1809.09600}}
  (\bibinfo{year}{2018}).
\newblock


\bibitem[\protect\citeauthoryear{Ying, Cai, Luo, Zheng, Ke, He, Shen, and
  Liu}{Ying et~al\mbox{.}}{2021}]%
        {ying2021transformers}
\bibfield{author}{\bibinfo{person}{Chengxuan Ying}, \bibinfo{person}{Tianle
  Cai}, \bibinfo{person}{Shengjie Luo}, \bibinfo{person}{Shuxin Zheng},
  \bibinfo{person}{Guolin Ke}, \bibinfo{person}{Di He},
  \bibinfo{person}{Yanming Shen}, {and} \bibinfo{person}{Tie-Yan Liu}.}
  \bibinfo{year}{2021}\natexlab{}.
\newblock \showarticletitle{Do Transformers Really Perform Bad for Graph
  Representation?}
\newblock \bibinfo{journal}{\emph{arXiv preprint arXiv:2106.05234}}
  (\bibinfo{year}{2021}).
\newblock


\bibitem[\protect\citeauthoryear{Yu, Dohan, Luong, Zhao, Chen, Norouzi, and
  Le}{Yu et~al\mbox{.}}{2018}]%
        {yu2018qanet}
\bibfield{author}{\bibinfo{person}{Adams~Wei Yu}, \bibinfo{person}{David
  Dohan}, \bibinfo{person}{Minh-Thang Luong}, \bibinfo{person}{Rui Zhao},
  \bibinfo{person}{Kai Chen}, \bibinfo{person}{Mohammad Norouzi}, {and}
  \bibinfo{person}{Quoc~V Le}.} \bibinfo{year}{2018}\natexlab{}.
\newblock \showarticletitle{Qanet: Combining local convolution with global
  self-attention for reading comprehension}.
\newblock \bibinfo{journal}{\emph{arXiv preprint arXiv:1804.09541}}
  (\bibinfo{year}{2018}).
\newblock


\bibitem[\protect\citeauthoryear{Yu, Jiang, Dong, and Feng}{Yu
  et~al\mbox{.}}{2019}]%
        {yu2019reclor}
\bibfield{author}{\bibinfo{person}{Weihao Yu}, \bibinfo{person}{Zihang Jiang},
  \bibinfo{person}{Yanfei Dong}, {and} \bibinfo{person}{Jiashi Feng}.}
  \bibinfo{year}{2019}\natexlab{}.
\newblock \showarticletitle{ReClor: A Reading Comprehension Dataset Requiring
  Logical Reasoning}. In \bibinfo{booktitle}{\emph{International Conference on
  Learning Representations (ICLR)}}.
\newblock


\bibitem[\protect\citeauthoryear{Zhang, Zhang, Xia, and Sun}{Zhang
  et~al\mbox{.}}{2020b}]%
        {zhang2020graph}
\bibfield{author}{\bibinfo{person}{Jiawei Zhang}, \bibinfo{person}{Haopeng
  Zhang}, \bibinfo{person}{Congying Xia}, {and} \bibinfo{person}{Li Sun}.}
  \bibinfo{year}{2020}\natexlab{b}.
\newblock \showarticletitle{Graph-bert: Only attention is needed for learning
  graph representations}.
\newblock \bibinfo{journal}{\emph{arXiv preprint arXiv:2001.05140}}
  (\bibinfo{year}{2020}).
\newblock


\bibitem[\protect\citeauthoryear{Zhang, Yang, Li, and Wang}{Zhang
  et~al\mbox{.}}{2019}]%
        {zhang2019machine}
\bibfield{author}{\bibinfo{person}{Xin Zhang}, \bibinfo{person}{An Yang},
  \bibinfo{person}{Sujian Li}, {and} \bibinfo{person}{Yizhong Wang}.}
  \bibinfo{year}{2019}\natexlab{}.
\newblock \showarticletitle{Machine reading comprehension: a literature
  review}.
\newblock \bibinfo{journal}{\emph{arXiv preprint arXiv:1907.01686}}
  (\bibinfo{year}{2019}).
\newblock


\bibitem[\protect\citeauthoryear{Zhang, Wu, Zhou, Duan, Zhao, and Wang}{Zhang
  et~al\mbox{.}}{2020a}]%
        {zhang2020sg}
\bibfield{author}{\bibinfo{person}{Zhuosheng Zhang}, \bibinfo{person}{Yuwei
  Wu}, \bibinfo{person}{Junru Zhou}, \bibinfo{person}{Sufeng Duan},
  \bibinfo{person}{Hai Zhao}, {and} \bibinfo{person}{Rui Wang}.}
  \bibinfo{year}{2020}\natexlab{a}.
\newblock \showarticletitle{SG-Net: Syntax-guided machine reading
  comprehension}. In \bibinfo{booktitle}{\emph{Proceedings of the AAAI
  Conference on Artificial Intelligence (AAAI)}}, Vol.~\bibinfo{volume}{34}.
  \bibinfo{pages}{9636--9643}.
\newblock


\bibitem[\protect\citeauthoryear{Zhang, Yang, and Zhao}{Zhang
  et~al\mbox{.}}{2021}]%
        {zhang2021retrospective}
\bibfield{author}{\bibinfo{person}{Zhuosheng Zhang}, \bibinfo{person}{Junjie
  Yang}, {and} \bibinfo{person}{Hai Zhao}.} \bibinfo{year}{2021}\natexlab{}.
\newblock \showarticletitle{Retrospective Reader for Machine Reading
  Comprehension}. In \bibinfo{booktitle}{\emph{Proceedings of the AAAI
  Conference on Artificial Intelligence (AAAI)}}, Vol.~\bibinfo{volume}{35}.
  \bibinfo{pages}{14506--14514}.
\newblock


\end{thebibliography}


\end{document}